\documentclass[journal]{IEEEtran}
\usepackage[utf8]{inputenc}
\usepackage{cite}
\usepackage{comment}
\usepackage{amsmath,amssymb,amsfonts}
\usepackage{booktabs}
\usepackage{color, colortbl}
\definecolor{Gray}{gray}{0.9}
\usepackage{graphicx}
\usepackage{xspace}
\graphicspath{{figs/}{bio-photos/}}
\usepackage[caption=false, font=footnotesize]{subfig}
\usepackage{multirow}
\usepackage{textcomp}
\usepackage[binary-units=true]{siunitx}
\usepackage{xspace}
\usepackage{float}

\newcommand{\rebuttal}[1]{{\color{black}#1}}

\newcommand{\figref}[1]{Figure~\ref{#1}}
\newcommand{\tabref}[1]{Table~\ref{#1}}
\newcommand{\secref}[1]{Section~\ref{#1}}

\newcommand{\fannOnMcu}{FANN-on-MCU\xspace}
\newcommand{\riscy}{RI5CY\xspace}
\newcommand{\ibex}{IBEX\xspace}

\usepackage{threeparttable}
\usepackage{amsmath}
\usepackage{upgreek}

\usepackage{url}

\usepackage{fancyhdr}
\fancypagestyle{mahmood}{%
  \fancyhf{} 
  
  \fancyhead[C]{\footnotesize \textcopyright 2020 IEEE. Personal use of this material is permitted.  Permission from IEEE must be obtained for all other uses, in any current or future media, including reprinting/republishing this material for advertising or promotional purposes, creating new collective works, for resale or redistribution to servers or lists, or reuse of any copyrighted component of this work in other works.}
}%
\makeatletter
\let\ps@IEEEtitlepagestyle\ps@mahmood
\makeatother

\begin{document}
\bstctlcite{IEEEexample:BSTcontrol}
%
\title{\fannOnMcu: An Open-Source Toolkit for Energy-Efficient Neural Network Inference at the Edge of the Internet of Things}

\author{Xiaying~Wang,~\IEEEmembership{Student Member,~IEEE,}
        Michele~Magno,~\IEEEmembership{Senior Member,~IEEE,}\\
        Lukas~Cavigelli,~\IEEEmembership{Member,~IEEE,}
        Luca~Benini,~\IEEEmembership{Fellow,~IEEE}
\thanks{Manuscript received November 07, 2019; revised January 16, 2020; accepted February 10, 2020.}
\thanks{The authors would like to thank \textit{armasuisse Science \& Technology} for funding this research. This project was supported in part by the Swiss Data Science Center (SDSC) PhD Fellowship under grant ID P18-04.}
\thanks{X. Wang, M. Magno, L. Cavigelli, and L. Benini are with the Integrated Systems Laboratory, ETH Zürich, 8092 Zürich, Switzerland (e-mail: xiaywang@iis.ee.ethz.ch). L. Benini is also with the Department of Electrical, Electronic and Information Engineering, University of Bologna, 40136 Bologna, Italy.} 
\thanks{Copyright (c) 2020 IEEE. Personal use of this material is permitted. However, permission to use this material for any other purposes must be obtained from the IEEE by sending a request to pubs-permissions@ieee.org.}
}

\maketitle{}

\begin{abstract}
The growing number of low-power smart devices in the Internet of Things is coupled with the concept of “Edge Computing”, that is moving some of the intelligence, especially machine learning, towards the edge of the network.  Enabling machine learning algorithms to run on resource-constrained hardware, typically on low-power smart devices, is challenging in terms of hardware (optimized and energy-efficient integrated circuits), algorithmic and firmware implementations. This paper presents FANN-on-MCU, an open-source toolkit built upon the Fast Artificial Neural Network (FANN) library to run lightweight and energy-efficient neural networks on microcontrollers based on both the ARM Cortex-M series and the novel RISC-V-based Parallel Ultra-Low-Power (PULP) platform. \rebuttal{The toolkit takes multi-layer perceptrons trained with FANN and generates code targeted to low-power microcontrollers.
This paper also presents detailed analyses of energy efficiency across the different cores, and the optimizations to handle different network sizes. Moreover, it provides a detailed analysis of parallel speedups and degradations due to parallelization overhead and memory transfers.} Further evaluations include experimental results for three different applications using a self-sustainable wearable multi-sensor bracelet. Experimental results show a measured latency in the order of only a few microseconds and power consumption of a few milliwatts while keeping the memory requirements below the limitations of the targeted microcontrollers.  In particular, the parallel implementation on the octa-core RISC-V platform reaches a speedup of 22x and a 69\% reduction in energy consumption with respect to a single-core implementation on Cortex-M4 for continuous real-time classification.
\end{abstract}


\IEEEpeerreviewmaketitle{}

\section{Introduction}
\IEEEPARstart{M}{achine learning} has been introduced into many tasks related to the Internet of Things (IoT) and mobile applications to address the major challenge of extracting relevant information from many sensors and data spread in the physical world.
Because of its high efficiency in extracting actionable information from large amounts of noisy raw data, machine learning will play a critical role in future IoT devices and services. Recent results on machine learning models demonstrate impressive classification accuracy, in some cases even outperforming humans \cite{HePReLU2015}. One essential feature of machine learning and in particular neural networks (NNs) is their flexibility that makes them suitable for a wide range of applications, including computer vision \cite{yoshimura2019deep,Cavigelli2015
}, natural language processing \cite{ray2018rise,Conti2018
}, biomedical \cite{Wang2018,zazzaro2019eeg}, and several others \cite{zanella2014internet,Wang2019,jeon2018ble,chianese2015smart}. 

Today, machine learning on IoT devices is applied with the traditional cloud computing paradigm where the whole data processing is performed in the cloud, and the IoT devices stream the data out in raw form or possibly after simple filtering and/or compression \cite{yoshimura2014analysis, samie2019cloud,Cavigelli2018a}. However, the number of IoT devices is expanding rapidly, and the massive amount of collected data is hard to manage by central clouds. The reasons are the massive workload on the IoT network, the cost of the communication infrastructure, the required energy for data transmission, and, more generally, reliability, latency, and privacy concerns \cite{premsankar2018edge,Andri2018}. 
The new trend of IoT devices is to be ``smart'' to make decisions on their own, without streaming all the raw data to the cloud \cite{yao2018msml}. The edge computing paradigm is pushing the data processing to the edge of the IoT (comprising gateways and embedded end-devices) close to the sensors where the data is collected \cite{samie2019cloud}. In many IoT applications, the computation can be distributed on different layers. For instance, the IoT device may perform the pre-processing of the data and transmit the intermediate results to the fog where the rest of the processing is performed \cite{qin2018power,Kaewkannate2016,Rault2017}.

This new generation of IoT devices is supplied by small-size batteries, which limits the available energy and computational resources. Thus, bringing intelligence to the edge is creating fascinating challenges for industrial and academic researchers \cite{
Cavigelli2017,
jeon2018ble,yoshimura2014analysis}. Lots of research efforts towards specialized hardware \cite{Andri2016,
Cavigelli2015a,
Bahou2018
} and optimized inference algorithms \cite{
Cavigelli2018} 
to run such NNs on power-constrained devices have been made over the last few years. Today's IoT devices host microcontrollers (MCUs), especially from the ARM Cortex-M family, which are able to achieve power consumption in the order of mW and computational resources in the order of hundreds of MOPS \cite{Rahimi2016,HePReLU2015,Magno2017a
}. The power consumption in the range of milliwatts is required for battery-operated devices to avoid frequent battery recharges. On the other hand, the computational resources of MCUs are often unable to perform on-board processing for complex algorithms and sensors for several application scenarios \cite{qin2018power,zazzaro2019eeg}. This results in very few examples of NNs that are running on milliwatt-powered MCUs, which are the most common compute engines available at the edge of the IoT~\cite{Rathore2016,Mayer2019,Wang2018,Murali2015,infiniwolf
}. To attenuate these computational limits, researchers are proposing new processing units to match the requirements of computational resources required by on-board data processing with state-of-the-art machine learning algorithms. The most promising approaches to improve the performance of ultra-low-power processors are parallelism, low-power fixed-function hardware accelerators \cite{Cavigelli2016,
Andri2019
}, memory access optimization \cite{Cavigelli2019},
and near-threshold technology \cite{Gautschi2017,
Andri2018a
}. Parallel architectures for near-threshold operation, based on multi-core clusters, have been explored in recent years with different application workloads \cite{Gautschi2017
} and low-power systems \cite{eggimann2019risc}.

On the software side, the majority of approaches in cloud computing are using deep convolutional NNs, which are incredibly accurate and well-suited to classify image frames but require a massive amount of memory and computational resources. However, there are many application scenarios, especially dealing with low-bandwidth time-series recordings from low-power sensors, where multi-layer fully-connected networks are just as effective \cite{yao2018msml,chen2018novel,Magno2017a,Nissen2003}. A well-known heuristic model is the multi-layer perceptron (MLP), a deep learning approach with multiple layers of interconnected intricate memory modules. The Fast Artificial Neural Network (FANN) library is an open-source neural network library \cite{Nissen2003}, which implements multi-layer artificial neural networks (ANNs) in C. To push FANN to its best in terms of energy efficiency on MCUs, it is essential that the implementation of the MLP model is optimized for the hardware architecture of the processors exploiting special instructions, parallelism, scratchpad memories, and hardware accelerators.

This paper presents FANN-on-MCU: an open-source framework for easy deployment of NNs trained with the FANN library on both ARM Cortex-M cores and new parallel ultra-low-power (PULP) RISC-V-based processors \cite{eggimann2019risc}. The former is the dominant processor core present in most MCUs, while the latter represents the forefront of open-source multi-core processors implementing the RISC-V instruction set architecture (ISA) and many custom instruction set extensions achieving high energy-efficiency and featuring widely-tunable performance for ultra-low-power embedded systems. Our framework offers automated deployments on MCUs with and without a floating-point unit, i.e., ARM Cortex-M0+/M4F and PULP-based processors such as Mr. Wolf or GAP8 \cite{ARM2017, pullini2018}. 

We present both the framework and the tools to train ANNs as well as detailed performance measurements on both ARM Cortex-M and PULP processors \rebuttal{and their comparison when varying the computational complexity and the memory footprint of ANNs.
Moreover, we evaluate the performance of FANN-on-MCU on a designed wearable system, which includes both an ARM Cortex-M4 core and a PULP-based processor, with three real-world applications using three different network sizes. 
Experimental results demonstrate that a parallel implementation on the PULP processor reaches up to 22$\times$ runtime speedup with a 69\% reduction in energy consumption with respect to a single core implementation on Cortex-M4 for continuous real-time classifications.}

The main contributions of the paper are:
\begin{itemize}
    \rebuttal{
    \item We present FANN-on-MCU: an open-source framework based on FANN Library. The framework allows building optimized multi-layer artificial neural network-based classifiers on all ARM Cortex-M family processors both with (i.e., ARM Cortex-M4F and M7F) and without a floating-point unit (i.e., ARM Cortex M0-M3), and on the very novel class of PULP processors based on the open-source RISC-V instruction set.
    }
    \item We provide extensive measurements and comparisons of the performance of our framework on both ARM Cortex-M and PULP processors implementing NNs of variable sizes, taking into account the memory footprint. \rebuttal{Such a detailed performance analysis has not been presented before in literature.
    \item We demonstrate optimizations to handle different sizes of networks considering the memory hierarchy of the underlying embedded processor.
    \item We provide detailed analysis of parallel speedups and degradations due to parallelization overhead and memory transfers where a parallel ultra-low power processor, such as the PULP-based Mr. Wolf, is used in an embedded system.
    }
    \item We implement fully connected networks for real-world application scenarios such as hand gesture recognition, fall detection for elderly people, and human activity classification using a multi-sensor wearable device based on an ARM Cortex-M4 MCU and PULP processor.
    \item We present experimental results with in-field measurements of memory usage, accuracy, feasible network sizes, and power consumption.
    \item We have released FANN-on-MCU as open-source software\footnote{Available at \url{https://github.com/pulp-platform/fann-on-mcu}} to help engineers and academics to have a powerful and easy-to-use tool to deploy ANNs on ultra-low-power IoT devices. 
\end{itemize}
We have organized the remainder of this paper as follows. 
In \secref{sec:relWork}, we shortly summarize the concept of MLPs, the FANN library, and FANNTool, and provide an overview of related work on deploying NNs at the edge. In \secref{sec:processors}, we provide an overview of our target platforms, the ARM Cortex-M series, and the RISC-V-based PULP series MCUs, together with a description of an application testbed we designed, named InfiniWolf, which includes both an ARM Cortex-M processor and a PULP processor. Then in \secref{sec:fannonmcu}, we describe our automated deployment toolkit and highly-optimized implementation for ARM Cortex-M targets and particularly also PULP-based systems. We extensively evaluate its performance in \secref{sec:perf_evaluation} on the supported MCU families and show results for several application showcases in \secref{sec:application} using our designed wearable platform. \rebuttal{We discuss the limitations and future work in \secref{sec:discussion}} before concluding the paper in \secref{sec:conclusions}.

\section{Related Work} \label{sec:relWork}

In the following subsections, we first provide a summary of MLPs and introduce the FANN library and FANNTool to train the networks which we then automatically deploy on different MCUs. Subsequently, we discuss currently available frameworks and libraries for deploying NNs on MCUs.

\subsection{Multi-Layer Perceptrons}

\begin{figure}[!t]
    \centering
    \subfloat[]{\includegraphics[trim={0.3cm 3.8cm 19.5cm 3.8cm}, clip=true, width=0.44\linewidth]{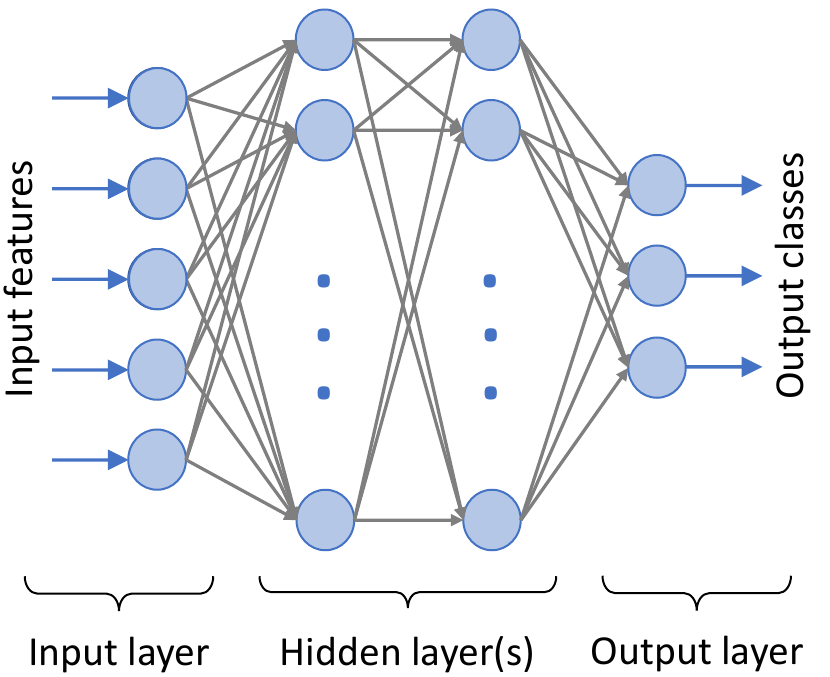}
    \label{fig:mlp_network}}
    \hfill
    \subfloat[]{\includegraphics[trim={16.8cm 4cm 0.3cm 4cm}, clip=true, width=0.526\linewidth]{mlp.pdf}
    \label{fig:mlp_neuron}}
    \caption{Multi-layer perceptron architecture (a) and building block (b).}
    \label{fig:mlp}
\end{figure}

A multi-layer perceptron is a type of feed-forward ANN, as illustrated in \figref{fig:mlp_network}, which consists of three or more layers of nodes: an input layer, one or more hidden layers, and an output layer. Each layer contains a fixed number of nodes, also called perceptrons or neurons, which (except for the input nodes) compute a weighted sum of the previous layer’s nodes and a bias, followed by a non-linearity such as sigmoid or ReLU functions. An illustration of the artificial neuron can be seen in \figref{fig:mlp_neuron}. In mathematical terms, each node computes
\begin{equation}
\centering
x_k^{(\ell+1)}(\textbf{x}) = \sigma \left(\sum_{i=1}^{m} w_{k,i}^{(\ell)}x_i^{(\ell)} + b_k\right),
\label{eq:mlp}
\end{equation}
where $x_k^{(\ell)}$ is the output of $k^\mathrm{th}$ node in $\ell^\mathrm{th}$ layer, $x_i$ are the elements of the input vector $\textbf{x}$ with dimension $m$, $w_{k,i}^{(\ell)}$ is the weight of the $i^\mathrm{th}$ input element to the node, and finally, $\sigma$ is the activation function.

The entire network is typically trained end-to-end by optimizing its parameters (weights and biases) using backpropagation and stochastic gradient descent, such that it maps the samples from the dataset to the labels to the best of its abilities (measured by a loss function). The MLP is specified by the number of hidden layers, the number of nodes within each layer, and which non-linearities are used (e.g., sigmoid).

MLPs are commonly used as the classifier after applying suitable non-learnable feature extractors due to their relatively moderately resource demand \cite{Magno2017}. In fact, MLPs are much less demanding than deep convolutional networks in terms of memory and computing requirements, and yet are widely used and very effective in many application areas such as medical data analysis and in general applications where hand-engineered feature extractors are preferred for interpretability~\cite{Dong2017}.

\subsection{Fast Artificial Neural Network (FANN) Library and FANNTool}
FANN \cite{Nissen2003} is an easy-to-use, mature, and well-documented framework to train and perform inference on multi-layer perceptrons. It is all written in C and has bindings to many languages, such as Python, MATLAB, Rust, etc. Due to its popularity, many graphical tools to aid training ANNs and selecting the right architecture and hyperparameters have become available. FANN also includes an automatic hyperparameter tuner and can optimize ANNs for fixed-point inference. \rebuttal{The library provides cascade training which automatically optimizes the number of hidden layers and neurons in an ANN, by starting with an empty neural network and then adding neurons one by one, while it trains the neural network.~\cite{Nissen2003}.}

FANN uses a simple file format for storing the dataset and the trained ANN model. It does not support training on GPUs, which simplifies the framework and is not required for networks of a size range suitable for deployment on low-power MCUs \cite{fanncortexm}. The CPU implementation is highly optimized with features such as cache optimizations and approximations of various activation functions as step-linear functions out-of-the-box. 

\rebuttal{To facilitate the usage of the library, the FANN community provides the FANNTool~\cite{Kuyumcu2007}. Its graphical interface allows for easy modification of the network architecture, activation function, training method, weight initialization, and monitoring of the training progress. It further supports the fully-automated selection of the network’s hyperparameters by iteratively testing all the available options present in FANN~\cite{Kuyumcu2007}.}

\subsection{Frameworks for Deploying NNs on MCUs}
With the growing attention to NNs, many frameworks have been developed over the last few years, such as PyTorch, TensorFlow, and Caffe2, with a focus on NNs training and cloud-scale deployment in GPU-accelerated data centers. However, only very recently, some focus has been put on low-power edge inference on MCUs. To the best of our knowledge, there are only two frameworks that are effectively working and have reached popularity: \emph{TensorFlow Lite for Microcontrollers} and ST Microelectronics' \emph{STM32Cube.AI}. 
\begin{itemize}
    \item \emph{STM32Cube.AI} can take trained models from Keras, TensorFlow Lite, and others to generate optimized code to run them on a wide range of MCUs of the STM32 series \cite{stmCubeAi2019}. For Keras, it also supports quantized models to reduce the model size and speed up the computation. 
    \item \emph{TensorFlow Lite for Microcontrollers} is not vendor-locked and much more generally supports platforms based on ARM Cortex-M and RISC-V~\cite{Louis2019}. It can export models from TensorFlow, its runtime takes up 16\,kB on a Cortex-M3, and it comes with implementations for floating-point layers as well as 8\,bit weights and activations with 32\,bit accumulation. 
\end{itemize}

These two frameworks highlight the commercial interest in deploying NNs on milliwatt-range edge devices. In contrast with those two commercial tools, the toolkit proposed in this paper provides a solution to deploy optimized MLPs on both ARM Cortex-M family (not limited to STMicroelectronics chips as with Cube.AI) and the PULP family, which is the leading-edge processor family based on RISC-V ISA. With respect to TensorFlow Lite for Microcontrollers, which is still under development, our toolkit supports parallel processing of ANNs on RISC-V-based processors and provides optimized code for the deployment of MCU systems with and without a floating-point unit. 

In parallel, ARM has been developing \emph{CMSIS-NN} \cite{Lai2018a}, an additional library for their \emph{Cortex Microcontroller Software Interface Standard} (CMSIS), in order to provide optimal-performance implementations of layers commonly present in deep NNs with speedups in the order of 4.6$\times$ over a simple baseline implementation. Our toolkit for the ARM Cortex-M family works on top of the CMSIS library to generate highly optimized code.

\fannOnMcu, the toolkit proposed in this paper, has been developed as an extension of \emph{FANNCortexM}~\cite{fanncortexm}. The latter first introduced the deployment flow shown herein, taking a trained MLP from FANN and exporting the code runnable on ARM Cortex-M MCUs with only minutes of engineering effort.
In this work, we optimize the implementation, provide much more detailed measurement results, introduce support for fixed-point models, and extended it with an optimized backend for deployment on the RISC-V-based PULP platform processors.
Moreover, this paper shows the benefits of parallel processing for ANNs evaluating the performance with three different application scenarios implemented in a working prototype.

\section{Low-Power Processors} \label{sec:processors}
As we mentioned in the previous section, one of the main contributions of this paper is the FANN-on-MCU toolkit, which enables MLP on low power processors, in particular MCUs. MCUs are the backbone of the majority of low-power smart devices for the IoT. In the following, we provide a short overview of the ARM Cortex-M family of MCUs, and we introduce the Mr. Wolf System on Chip (SoC) we use as a representative of the RISC-V-based family of PULP processors. Both processors have a power consumption of milliwatts and are suitable for small-size (hundreds of mAh) battery operation. Moreover, we describe InfiniWolf, an embedded platform we designed that incorporates both an ARM Cortex-M and a PULP processor to enable low-power real-time wearable applications.

\subsection{Low-Power Embedded Processing: ARM Cortex-M Family} 
The ARM Cortex-M (M0, M3, M4, M7) family features different computational capabilities and operating frequencies and has power consumption in the milliwatt range. The typical frequency is 16\,MHz for the M0 and up to 300\,MHz for the new and more powerful M7. The majority of those processors have no floating-point unit, and only the ARM Cortex-M4F and M7 implement a floating-point unit. This family of MCUs is characterized by an on-chip SRAM of a few kB (256kB-512kB) and a non-volatile flash memory with a maximum size of 1-2\,MB. The flash memory is typically used to preload the program code and static data, while the SRAM is used for the runtime code and main data memory. Thus, one of the constraints to be taken into account is the size of the non-volatile memory. Moreover, as MCUs are designed with low power and low cost in mind, the operations per second and a small memory footprint need to be taken into consideration. Finally, it is important to notice that some ARM Cortex-M cores have an integrated digital signal processing (DSP) instruction set. This is the case of all the ARM Cortex-M3, M4, and M7. The DSP instructions can be used to accelerate many of the operations used for signal processing and data analysis (e.g., the Fast Fourier transform). Most notably, the Cortex-M4 and Cortex-M7 have integrated single instruction, multiple data (SIMD) instructions, and multiply-and-accumulate operations (MACs) that might be exploited to accelerate computation in NNs. 

The Cortex Microcontroller Software Interface Standard (CMSIS) is a vendor-independent hardware abstraction layer for the Cortex-M processor series and defines generic tool interfaces \cite{ARM2017}. CMSIS enables consistent device support and simple software interfaces to the processor and the peripherals, simplifying software reuse, reducing the learning curve for MCU developers, and reducing the time to market for a new device. ANNs have a high number of multiplications, so minimizing the computation time increases the efficiency of the solution. We extensively use the optimized CMSIS floating-point and fixed-point multiplication function \texttt{arm\_dot\_prod}, where we measured a decrease of the execution time by 36\%, which shows the effectiveness of CMSIS. Among others, we have also used \texttt{arm\_fill} and \texttt{arm\_copy} that also gave an improvement in the range of 30\% in execution time. Moreover, CMSIS optimizes the computational time of many functions, such as DSP operations. The DSP library includes over 60 digital signal processing related functions that are optimized for the Cortex-M processors. DSP functions can be handy for feature extraction (i.e., to perform Fast Fourier transforms) but are not required for this toolkit, which can run with optimized performance also on ARM Cortex-M0 and other processors without DSP instructions.

\subsection{Parallel Ultra-Low-Power Platform: PULP Processors}

\begin{figure}[!t]
	\centering
	\includegraphics[trim={2cm 2cm 2cm 1.8cm},clip=true, width=\columnwidth]{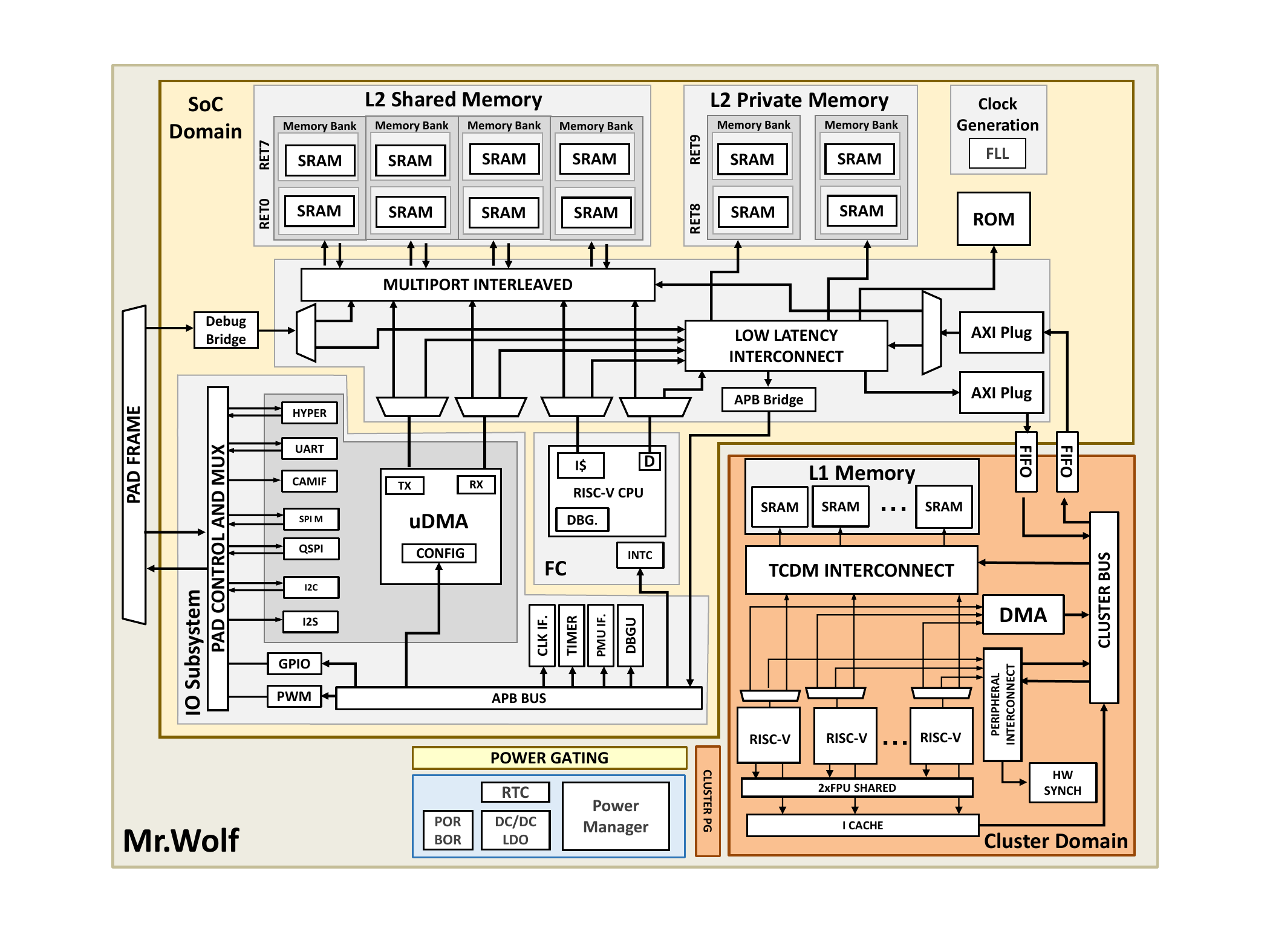}
	\caption{Block diagram of Mr. Wolf.}
	\label{fig:mrwolf-diagram}
\end{figure}

\begin{figure}[!t]
	\centering
	\includegraphics[trim={2.2cm, 2.5cm, 2.5cm, 5cm},clip=true, width=\linewidth]{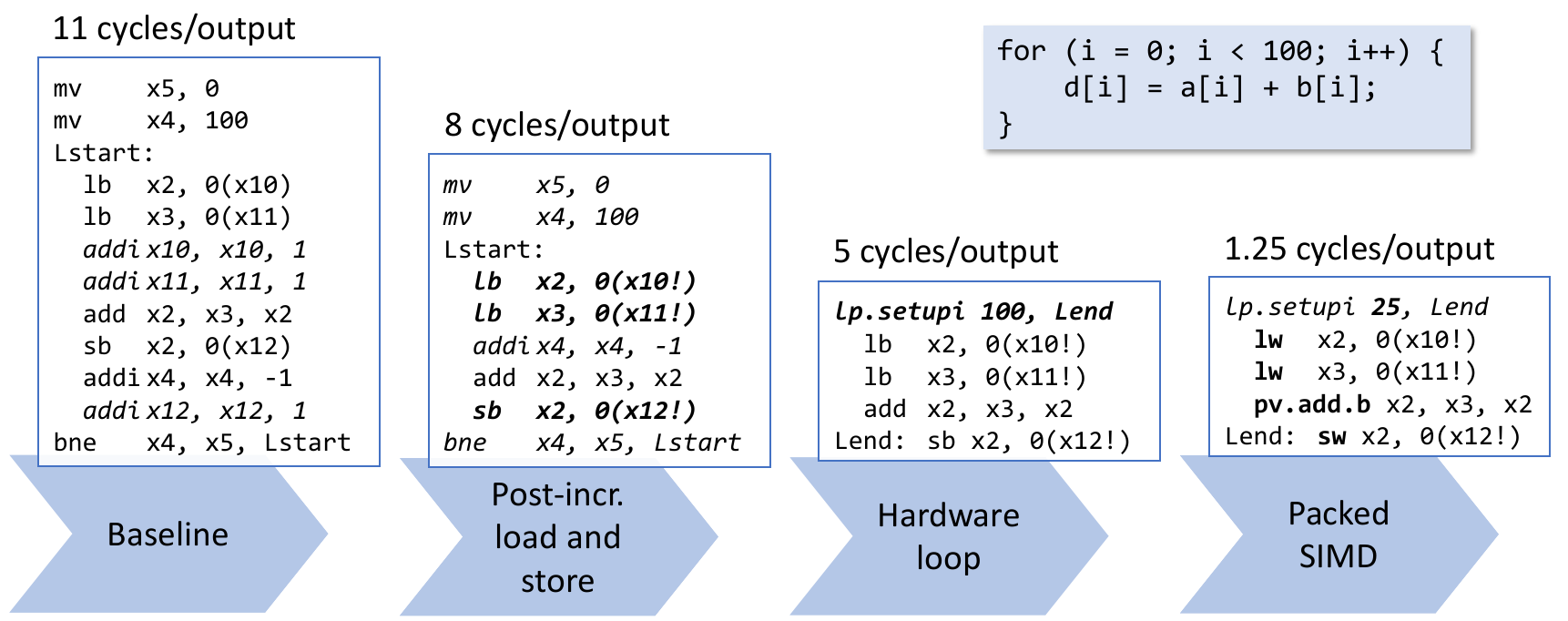}
	\caption{RISC-V ISA extensions of PULP.}
	\label{fig:pulp-isa-extensions}
\end{figure}

The PULP platform is an open-source, multi-core platform based on the RISC-V ISA achieving leading-edge energy-efficiency and featuring widely-tunable performance within a power envelope of a few mW \cite{eggimann2019risc}. PULP aims to satisfy the computational demands of IoT applications, which require flexible and fast processing of data streams generated by multiple sensors, such as accelerometers, microphone arrays, low-resolution cameras, vital signs monitoring sensors. As opposed to single-core MCUs, a parallel ultra-low-power programmable architecture provides the ability to meet the computational requirements of applications in IoT domains where low latency and low energy consumption are the central keys for solving tasks on miniaturized, battery-powered systems. Despite being an academic research platform, PULP offers the maturity of a commercial device with OpenMP, OpenCL, and OpenVX support to enable agile application porting, development, performance tuning, and debugging. GreenWaves Technologies produces commercial devices based on the open-source PULP platform to design ultra-low-power embedded solutions for image, sound, and vibration AI processing in sensing devices \cite{8715489}.

In this work, we chose PULP Mr. Wolf processor~\cite{pullini2018} for its ultra-low-power scalable performance, designed explicitly for always-on AI-powered IoT applications. We report its block diagram in \figref{fig:mrwolf-diagram}. Mr. Wolf features a hierarchical architecture with a small RISC-V core in the so-called fabric controller (FC) subsystem. It is coupled with an autonomous I/O subsystem for efficient data transfers from a broad range of peripherals exploiting a multi-channel I/O direct memory access ($\upmu$DMA) unit in the SoC domain. The small core can offload compute-intensive tasks to a parallel eight-core processing engine in the cluster domain, which is activated only on demand. 512\,kB of L2 memory is available in the SoC domain and it is divided into a shared L2 memory arranged in four 448\,kB memory banks for easy access from both $\upmu$DMA and processors, and a private L2 memory for the FC to store, for example, program, stack, private data, in order to minimize bank conflicts. On the other hand, the Cluster domain is equipped with a multi-bank L1 memory (sixteen 4\,kB SRAM banks) to serve the parallel access from the eight cores. The data transfer from/to the L2 memory to/from the L1 memory is handled by an autonomous DMA unit. The memory organization of the whole system is designed in such a way that the interaction and the access conflicts are extremely minimized.

The small core in the FC is called \ibex~\cite{Schiavone2017a} and implements the basic RV32IMC ISA, while the Cluster domain comprises eight \riscy cores with custom instruction set extensions for DSP including hardware loops, post-incremental load and store instructions and additional ALU instructions. \figref{fig:pulp-isa-extensions} shows the cycle reduction using the ISA extensions. We can see that with the post-incremental load and store and the hardware loop, we can gain a $2\times$ speedup with respect to the baseline RV32IMC ISA, additionally with packed SIMD instructions, we can achieve up to approximately 10$\times$ speedup. Finally, the operational frequency of the SoC can be configured between 32\,kHz and 450\,MHz, while for the cluster, the range is 32\,kHz--350\,MHz.

In this work, we fully exploit the memory hierarchy of PULP chips and the custom ISA extensions, i.e., the hardware loop and the post-incremental load and store instructions, and more significantly, we will apply the cluster parallelism to boost the computational capacity of the whole system.

\subsection{Testbed Platform: InfiniWolf}

\begin{figure}[!t]
	\centering
	\includegraphics[width=0.9\columnwidth]{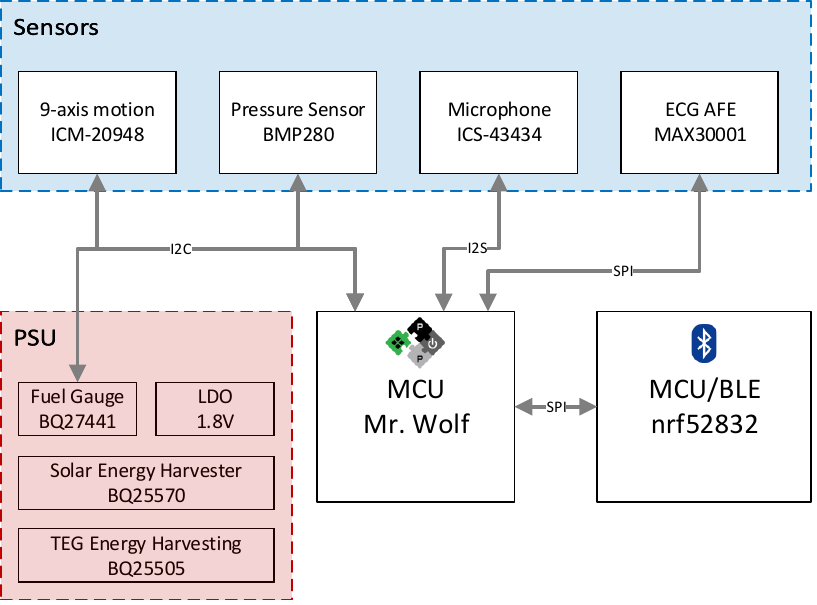}
	\caption{Block diagram of InfiniWolf and the smart power unit that is able to harvest energy from dual sources.}
	\label{fig:block_diagram}
\end{figure}

\begin{figure}[!t]
	\centering
	\includegraphics[width=0.75\columnwidth]{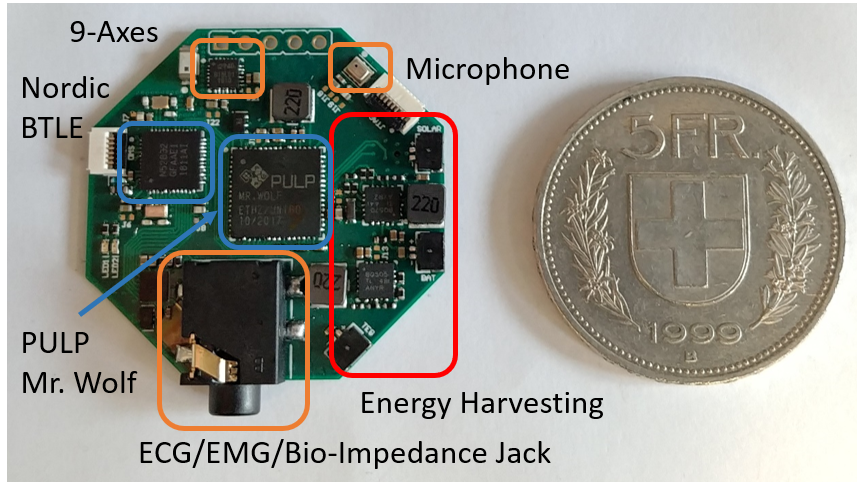}
	\caption{InfiniWolf prototype used to carry out experimental measurements.}
	\label{fig:board}
\end{figure}

To apply FANN-on-MCU to real-world applications, we designed a hardware platform and used it as a testbed. The designed platform is called InfiniWolf~\cite{infiniwolf}, a wearable battery operated multi-sensor device that aims to work perpetually with small-size energy harvesting and can be worn as a smartwatch.
\figref{fig:block_diagram} shows the block diagram and the architecture of the designed platform, which features two processors, a Nordic nRF52832 Bluetooth low energy SoC with an ARM Cortex-M4 processor and Mr. Wolf. The Nordic SoC handles communication with a remote host and offers auxiliary support for small data processing if needed. It provides Bluetooth Low Energy (BLE) 5 communication capabilities, performs power management in various modes of operation (sleep, raw data streaming, data acquisition, and processing), and keeps track of the battery charging status. 
The dual-processor architecture of InfiniWolf allows local end-to-end processing (i.e., on-board classification using machine learning) with lower power and higher energy efficiency than streaming the data out for remote analysis \cite{Palossi2018}. Moreover, this architecture allows lower latency in the range of $\upmu$s with respect to wireless connectivity.

\rebuttal{A dual-source energy harvester for solar and thermoelectric generator (TEG)~\cite{Thielen2017TEG} modules has been included in the design.} The main goal is to achieve perpetual work when the energy transducers are deployed on a wrist band harvesting energy from light and body heat. The choice of two energy harvesting sources is motivated by increased flexibility and robustness of the energy intake, while form-factor is not compromised as the two harvesters exploit different sides of the watch (top for solar and bottom for thermal). Assuming InfiniWolf staying in challenging indoor conditions for 6 hours, and the worst-case scenario for the TEG energy harvester, it will acquire 21.44\,J per day. 
\rebuttal{This energy can be used to prolong the battery lifetime of InfiniWolf or to achieve an energy-autonomous smart watch~\cite{Magno2016Infinitime}. In the latter case, the energy acquired needs to balance the energy consumed during the classification and the power consumption for the sleep mode. Thus, it is crucial both limiting the activation time and the energy in active mode, as well as reducing the power consumption in sleep mode~\cite{Magno2016Infinitime}.}
The smart power supply unit (PSU) includes a TEG energy harvesting integrated circuits based on the BQ25505 from Texas Instruments, while the BQ25570 integrated circuits deal with the solar energy harvesting and provide a 1.8 LDO. Finally, a fuel gauge integrated circuit (BQ27441) monitors the 120\,mAh Li-Ion battery. 
The smartwatch includes a 9-axis motion sensor (Invensense ICM20948), a pressure sensor (Bosch Sensortec BMP280), a microphone (Invensense ICS-43432), and an ultra-low-power ECG/EMG and bioimpedance analog front-end (Maxim MAX30001) to acquire biomedical signals as well as a low power galvanic skin response (GSR) front-end. The wearable device can be worn on the user's wrist and periodically and opportunistically acquires information from the sensors according to the available energy. 
A prototype of the smartwatch is shown in \figref{fig:board}.

\section{Open-Source \fannOnMcu Framework} \label{sec:fannonmcu}

This section provides an overview of the steps for developing an embedded machine learning system and illustrates \fannOnMcu in detail.

\subsection{Embedded Machine Learning System Development}
\begin{figure}
    \centering
	\includegraphics[trim={1.2cm, 3.35cm, 14cm, 1.5cm},clip=true, width=\linewidth]{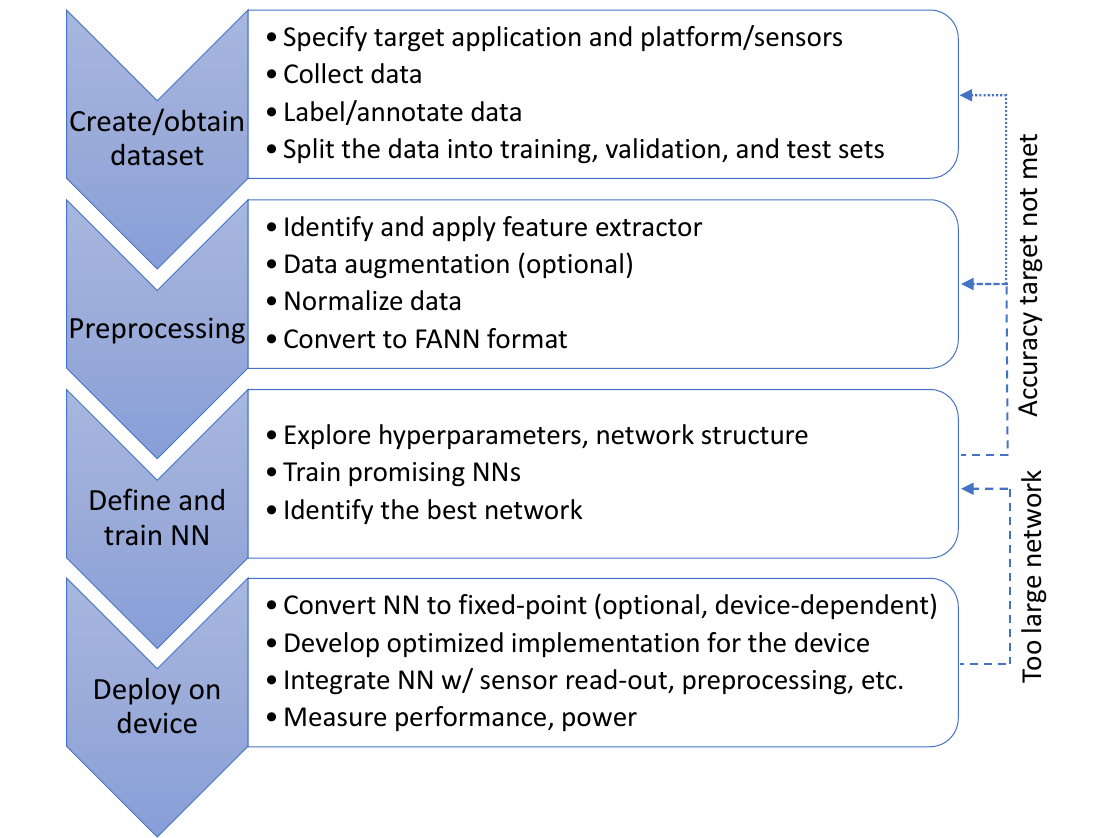}
	\caption{Process flow from concept to deployment.}
	\label{fig:fannWorkflow}
\end{figure}

Developing an intelligent sensor device with on-board processing capabilities encompasses many steps. We provide a brief overview in \figref{fig:fannWorkflow}. The development starts with specifying a precise target application and which platform and sensors to use, collecting data, and annotating it with the desired labels. In the next step, the data should be pre-processed, identifying and applying suitable feature extractors, optionally performing data augmentation, normalizing the data, and preparing it in a suitable data format for training. Then the neural network has to be specified and trained, the network’s hyper-parameters and its structure have to be explored (number of layers and nodes, which activations), promising networks have to be trained, and the best network has to be identified. If the desired accuracy cannot be obtained, the previous steps have to be revisited (collection of more data, different feature extractors, changing the data augmentation for the training samples). Once the desired accuracy has been achieved, the network must be deployed on the device, converting the neural network to fixed-point (if no hardware floating-point support is available), developing an optimized implementation for the target device, co-integrating it with the sensor read-out and pre-processing, and measuring the resulting performance and power. 

\subsection{The \fannOnMcu Deployment Framework}
The open-source \fannOnMcu framework allows multi-layer ANNs to be implemented and optimized on resource-constrained ultra-low-power MCUs \rebuttal{by calling a single-line of command. It automatically generates the optimized C code that can be directly compiled for the selected platform. In particular, the framework takes into account the platform architecture by exploiting the ARM M-series-specific instructions, and the custom XPULP instruction set extensions on PULP.} The framework includes a script for automatic conversion of the trained network to a directly callable dependency-free C function, including a test method that applies samples from the dataset for verification and benchmarking. Notably, the generated code files include all the parameters of the network to overcome the need for file system support. This allows for a straightforward workflow: 

\begin{enumerate}
    \item Convert the data to the standard FANN format;
    \item Train a neural network using FANN and save it; 
    \item Optionally convert the neural network to fixed-point by rescaling the input data and calling \texttt{fann\_save\_to\_fixed};
    \item Apply the conversion script to the ANN and dataset; 
    \item Call the \texttt{fann\_type* fann\_run(fann\_type *input)} function from within your code; 
    \item Build your code together with the generated files and evaluate the resulting application.
\end{enumerate}
Our converter also supports fixed-point and floating-point models and can thus run also efficiently on MCUs without a floating-point unit, such as the tiny Cortex-M0 and M0+. We have released the code as open-source software online].

The framework is straightforward to use: with a single-line command, one can generate the C code and header files for the desired platform with the selected data type, i.e., floating-point or fixed-point. The conversion script takes into account the processor family and evaluates the network size to automatically select the level of memory closest to the processing unit, still big enough to contain the whole network. Specifically, it estimates the required memory for the network and the buffers according to 
\begin{equation} \label{eq0}
\begin{split}
E_m = &(2 \cdot L_\mathrm{data\_buffer} + 5 \cdot N_\mathrm{neurons} + \\
 & N_\mathrm{weights} + 2 \cdot N_\mathrm{fann\_layers}) \cdot \mathrm{sizeof}(\mathrm{dtype}), 
\end{split}
\end{equation}
where $E_m$ is the estimated memory size, $L_\mathrm{data\_buffer}$ is the buffer length of one input sample multiplied by two considering the eventual double buffering for continuous data processing from sensors or other sources, $N_\mathrm{neurons}$ is the total number of neurons in the network including the biases seen as an additional neuron to every layer, this constant is multiplied by five due to the storage of indexes to the first and last connections in the layer, the activation steepness, the type of the activation function, and the output of every neuron, $N_\mathrm{weights}$ is the total number of the weights in the network, and $N_\mathrm{fann\_layers}$ is the total number of layers including the input, the hidden, and the output layers, multiplied by two for storing the indexes of the first and the last neurons in each layer.

According to $E_m$ and the selected processor, the framework automatically stores the network parameters in the level of cache that is closest to the processing unit and still contains the network. For example, if an ARM Cortex-M processor with 96\,kB RAM available for data storage is selected and the estimated network size is 30\,kB, the network parameters will be automatically loaded into RAM. Additionally, for PULP-based processors, the framework is aware of the characteristic double domain feature of some 
PULP-based processors and the DMA unit. For example, the following situations apply for PULP Mr. Wolf:
\begin{itemize}
    \item FC selected, $E_m$ smaller than the private L2 memory, then the network is stored into the private memory of the FC.
    \item FC selected, $E_m$ bigger than the private L2 memory, then the network is stored in the shared L2 memory.
    \item Cluster selected, $E_m$ smaller than the L1 memory, then the network is stored into L1 memory.
    \item Cluster selected, $E_m$ bigger than the L1 memory, then the network is stored into the shared L2 memory with automatic DMA transfers exploiting the double buffering, i.e., while computing on one chunk of data, the next chunk is transferred simultaneously. In this case, two types of transfers are possible:
    \begin{itemize}
        \item When the largest layer fits into the L1 memory, the DMA unit transfers the whole layer from L2 to L1 (i.e., layer-wise transfers).
        \item When the largest layer does not fit into the L1 memory, the DMA unit performs neuron-wise transfers, i.e., it transfers the weights for a single neuron at a time.
    \end{itemize}
\end{itemize}

Generally, in PULP-based processors, the cluster domain is used for applications where a large amount of computations is required, while the FC is usually engaged in I/O handling or other kinds of scheduling. However, for an application scenario where, for example, a small network is used to detect the onset and, once the onset is detected, a deeper network is used for classification~\cite{coninck2016}, both domains (SoC and Cluster) have its own advantage: the FC continuously reads the sensory data and executes the onset detection algorithm, while the cluster domain is activated once the onset is detected to perform the classification with a deep NN. In this case, our proposed framework stores the small network into the private L2 memory for the FC, while for the classification task, the DMA unit transfers the network using the double-buffering technique into the L1 memory for the computation in cluster domain. With this configuration, access to memory is the fastest for both computation domains. This way, we successfully meet the two main requirements in the IoT domain: low power and low latency, since we power on the powerful computation engine, i.e., the cluster domain, to perform fast computation only when it is strictly necessary in order to save the overall energy consumption. The whole process is automated in our framework to alleviate the user's workload.

\section{Performance Evaluation} \label{sec:perf_evaluation}
\begin{figure}[!t]
	\centering
	\includegraphics[trim={0.2cm, 0.2cm, 0.2cm, 0.68cm},clip=true, width=\columnwidth]{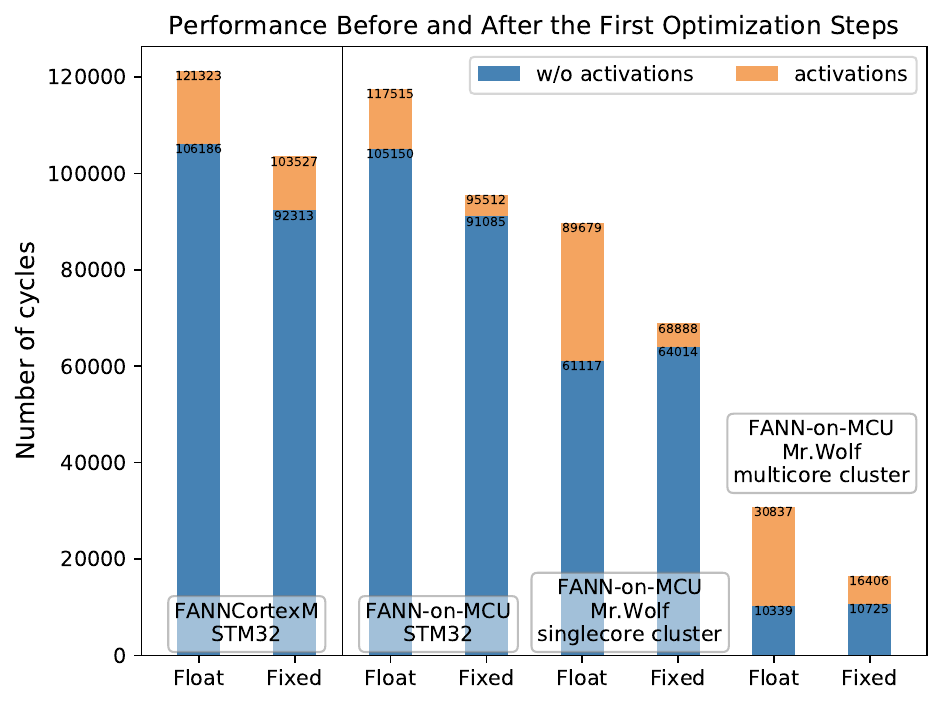}
	\caption{Runtime in number of cycles before and after the first optimization steps on the example network.}
	\label{fig:example_network}
\end{figure}

In this section, we evaluate the performance of \fannOnMcu on a wide variety of network architectures with varying input and output sizes and number of hidden layers and hidden units, before demonstrating its efficiency for specific application showcases in \secref{sec:application}.

\subsection{Methods}

\begin{figure*}[!t]
     \centering
     \subfloat[]{\includegraphics[trim={0.25cm 0.25cm 0.25cm 0.6cm}, clip=true, width=0.5\linewidth]{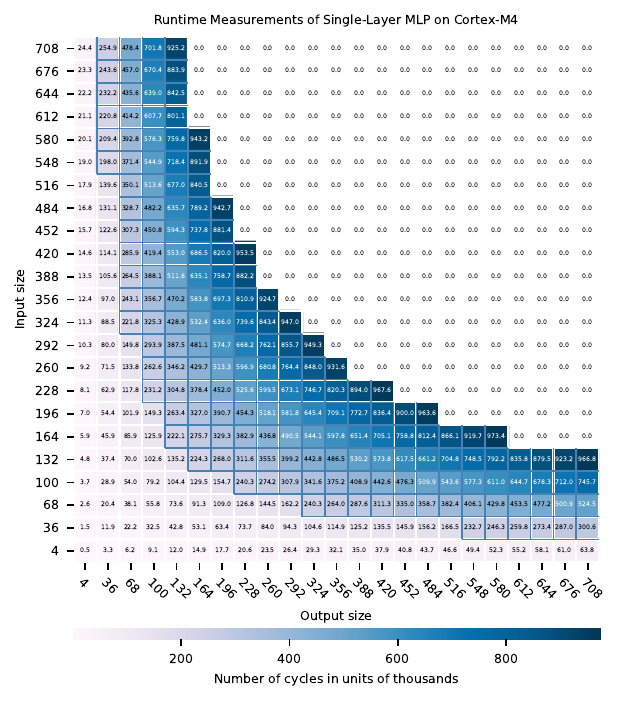}\label{fig:single_layer_arm}}
     \hfill
     \subfloat[]{\includegraphics[trim={0.25cm 0.25cm 0.25cm 0.6cm}, clip=true, width=0.5\linewidth]{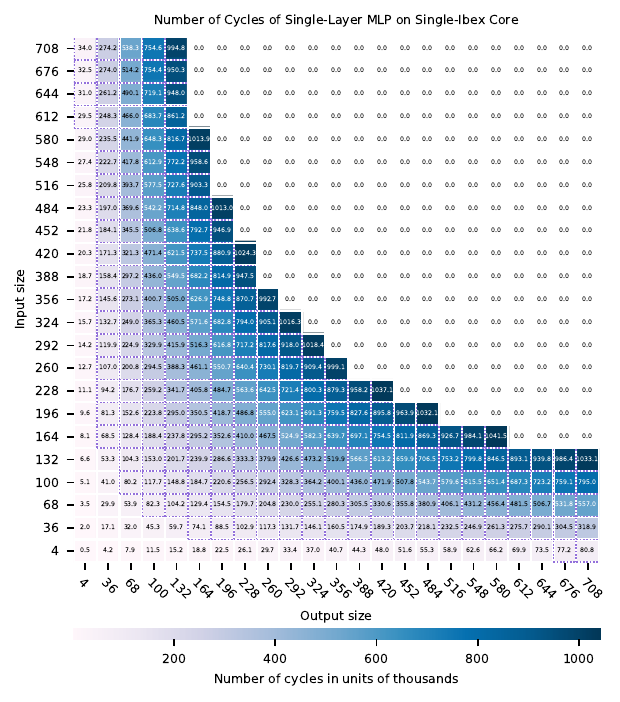}\label{fig:single_layer_fc}}
    \caption{Runtime in number of cycles of a single layer by varying the number of input and output to the layer. 0.0 is when the network is too big to be stored in the largest memory. Runtime measurements on (a) an ARM Cortex-M4 and (b) the PULP \ibex core with the basic RV32IMC ISA. In (a), the continuous blue grid delimits the case when the layer is too large to fit into RAM, therefore it is stored in flash memory. In (b), the purple dotted grid delimits the case when the layer is too large to fit into private L2, hence it is stored in the shared L2 memory.}
    \label{fig:num_cycles_arm_pulp_fc}
\end{figure*}

\begin{figure*}[!t]
     \centering
     \subfloat[]{\includegraphics[trim={0.25cm 0.25cm 0.25cm 0.6cm}, clip=true, width=0.5\linewidth]{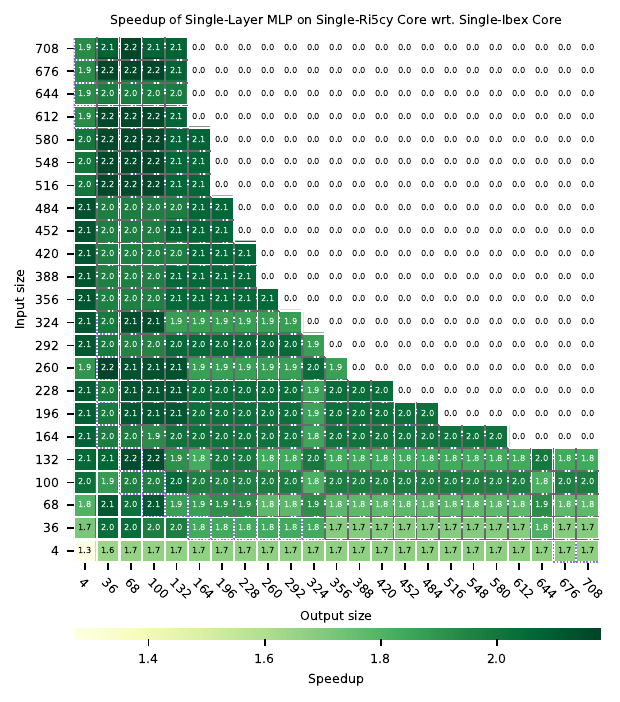}\label{fig:single_layer_single_riscy_fc}}
     \hfill
     \subfloat[]{\includegraphics[trim={0.25cm 0.25cm 0.25cm 0.6cm}, clip=true, width=0.5\linewidth]{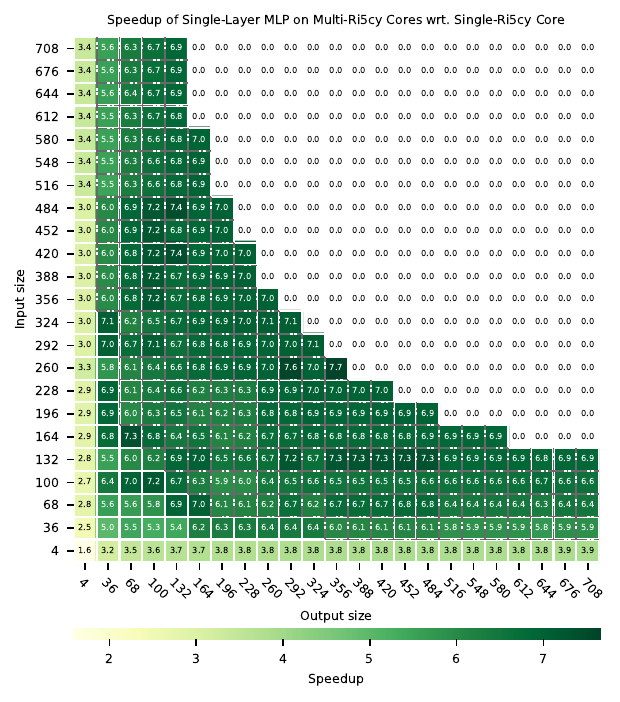}\label{fig:single_layer_multi_riscy_single_riscy}}
    \caption{Speedup on PULP of a single layer perceptron by varying the number of input and output to the layer. The purple dotted grid delimits the case when the layer is too large to fit into private L2, hence it is stored in the shared L2 memory for the FC. The gray dash-dotted grid delimits the case when the layer does not fit into the L1 memory and neuron-wise DMA transfer is applied for the Cluster. 0.0 is when the layer is too big to be stored in the largest memory. (a) Speedup of single \riscy core with respect to a single \ibex core, (b) Parallel speedup of multiple \riscy cores with respect to a single one.}
    \label{fig:speedup_pulp}
\end{figure*}

\rebuttal{In a first step, we analyzed the code provided by \emph{FANNCortexM}, which supports only ARM Cortex-M, to find performance bottlenecks and to do profiling using an example network with 5 input features, 2 hidden layers with 100 neurons each, and 3 output classes. Each neuron is followed by an hyperbolic tangent as activation function. The evaluation is done on an STM32L475VG with ARM Cortex-M4. Moreover, we run the same network on Mr. Wolf to show comparisons in both floating and fixed point.}

We then proceeded with the implementation of \emph{\fannOnMcu} to support the deployment of ANNs on both ARM Cortex-M series and PULP family, and exhaustively evaluated its performance on STM32L475VG with ARM Cortex-M4 and PULP-based Mr. Wolf with the following two approaches:
\begin{itemize} 
    \item We first considered single-layer performance. We measured the runtime of a single layer by varying the number of inputs and outputs of the layer, namely the length of the input feature vector and the number of neurons in the layer.
    \item Subsequently, we varied the number of hidden layers with fixed input features (100 input features) and fixed output units (8 classes) and measured the performance of the whole network. The number of hidden units in every hidden layer varies according to the following equation:
    
    \begin{equation} \label{eq1}
    N_l = (l\ \mathbf{mod}\ 2 + l\ \mathbf{div}\ 2) \cdot d
    \end{equation}
    
    where $N_l$ is the number of neurons in $l$-th hidden layer, $\mathbf{mod}$ is the modulo operation and $\mathbf{div}$ is the integer division, $d$ a tunable parameter. Hence, the number of total hidden units ($N_{tot}$) is:
    
    \begin{equation} \label{eq2}
    N_{tot} = \sum_{l=1}^{L} (l\ \mathbf{mod}\ 2 + l\ \mathbf{div}\ 2) \cdot d
    \end{equation}
    
    with L the total number of hidden layers. \rebuttal{The tunable parameter $d$ decides how fast the number of neurons in each layer increases, i.e. how fast the whole network grows before reaching the memory limitation of the embedded platform.}
\end{itemize}
\rebuttal{All the measurements are done on-board in number of cycles without any normalization. The results are shown and discussed in the next subsection.}

\subsection{Experimental Results and Discussion}

\rebuttal{The inference of MLP comprises of two main parts: it first computes the linear function in \eqref{eq:mlp} by calculating the inner dot-products, then it executes the non-linear activation function. We did our evaluations following this separation. We provide results for both floating-point and fixed-point implementations, as many Cortex-M and RISC-V devices, which are more cost-efficient, do not include a floating-point unit.

In the previous \emph{FANNCortexM}, the buffer used to keep the intermediate values of each neuron is first filled in with the bias and then overwritten immediately afterwards. This step causes an unnecessary performance slowdown, which is eliminated in our version of \emph{\fannOnMcu}.  

\figref{fig:example_network} shows the gain in performance and the profiling results of the example network executed on the STM32L475VG's ARM Cortex-M4. In the plot, we can observe three main outcomes on Cortex-M4: 1) The elimination of unnecessary initialization improves the runtime by 3.1\% and 7.7\% for floating-point and fixed-point implementations, respectively; 2) The fixed-point version is around 15\% faster than the floating-point version; 3) The computation of the weight matrices without the activation functions is the most computationally demanding part. More specifically, it comprises approximately 88\% of the total compute time for this example network. Based on these observations, we conducted our following analyses focusing on the computation of weight matrices without activation functions.
For Mr. Wolf, we observe 1) the same network running on a single \riscy core is around 1.3$\times$ and 1.4$\times$ faster than the Cortex-M4 in floating and fixed point, respectively; 2) The parallelization provides up to 6$\times$ speedup in both floating and fixed point.

These results are well aligned with the expectations: 
1) On the Cortex-M, the floating- and fixed-point implementations require 8 and 7 cycles in the inner-most loop (cf. \tabref{table:arm_pulp_instr}), respectively, which corresponds accurately to the ratio of the cycle counts reported in \figref{fig:example_network}; Similarly, the ratio of the cycle counts between the Cortex-M and single-core \riscy implementations match the expected 7/5 and 8/5 factors for fixed/float, respectively;
2) The floating-point and fixed-point implementations on \riscy both require 5 instructions, each taking 1 cycle, and thus resulting in similar execution times. Even after parallelization we do not see a significant performance difference between the two versions despite sharing only 2 floating-point units (FPUs) among the 8 cores. However, as only every fifth instruction uses the FPU, its utilization reaches 80\% and is thus not a performance bottleneck. 

}

\begin{table*}[!t]
\caption{\rebuttal{Assembly code for inner-loop dot-product. The number of cycles taken by the instruction is reported in parenthesis.}}
\label{table:arm_pulp_instr}
\centering
\begin{tabular}{@{}llll@{}}
\toprule
\begin{tabular}[c]{@{}l@{}}ARM Cortex-M4\\ Float\end{tabular}                                                                                                                & \begin{tabular}[c]{@{}l@{}}ARM Cortex-M4\\ Fixed\end{tabular}              & \begin{tabular}[c]{@{}l@{}}RISC-V RI5CY\\ Float\end{tabular}                                                                     & \begin{tabular}[c]{@{}l@{}}RISC-V RI5CY\\ Fixed\end{tabular}                                                                                   \\ \midrule
\begin{tabular}[c]{@{}l@{}}vldmia.32 (post-incr. load, 1)\\ vldmia.32\\ subs (counter subtract, 1)\\ vfma.f32 (fused mult-add, 3)\\ bne (taken conditional branch, 2)\end{tabular} & \begin{tabular}[c]{@{}l@{}}ldr (post-incr. load, 1)\\ ldr (post-incr. load, 1)\\ mul (multiply, 1)\\ add (addition, 1)\\ subs (counter subtract, 1)\\ bne (taken conditional branch, 2)\end{tabular} & \begin{tabular}[c]{@{}l@{}}flw (float load, 1)\\ flw\\ addi (pointer incr., 1)\\ addi\\ fmadd.s (fused mult-add, 1)\end{tabular} & \begin{tabular}[c]{@{}l@{}}p.lw (post-incr. load, 1)\\ p.lw\\ mul (multiply, 1)\\ sra (shift right arith., 1)\\ add (addition, 1)\end{tabular} \\
                                                                                                                           \midrule                                                  & 4$\times$ loopunrolling                                                           &                                                                                                                                  & 2$\times$ loopunrolling                                                                                                                               \\ \bottomrule
\end{tabular}
\end{table*}

\begin{figure*}[!t]
    \centering
    \subfloat[]{\includegraphics[trim={0.25cm 0.25cm 0.25cm 0.6cm}, clip=true, width=0.5\linewidth]{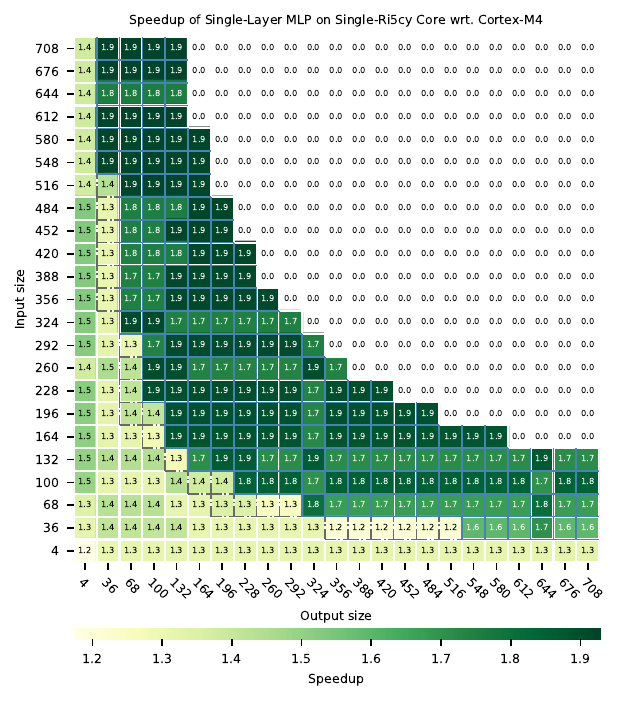}\label{fig:single_layer_arm_single_riscy}}
    \hfill
    \subfloat[]{\includegraphics[trim={0.25cm 0.25cm 0.25cm 0.6cm}, clip=true, width=0.5\linewidth]{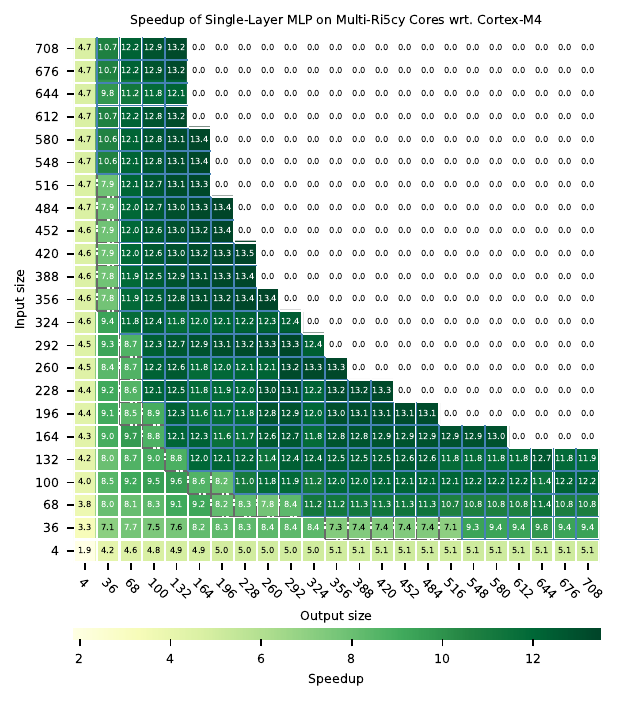}\label{fig:single_layer_arm_multi_riscy}}
    \caption{Comparison between ARM Cortex-M4 and PULP \riscy cores on a single layer perceptron by varying the number of input and output to the layer. The gray dash-dotted grid delimits the case when the layer is too big for the L1 memory and neuron-wise DMA transfer is applied for \riscy. The continuous blue grid delimits the case when the layer is too large to fit into RAM, hence it is stored in flash for Cortex-M4. 0.0 is when the layer is too big to be stored in the largest memory. (a) Speedup of single \riscy core with respect to ARM Cortex-M4, (b) Parallel speedup of multi \riscy with respect to ARM Cortex-M4.}
    \label{fig:speedup_comparison_arm_pulp}
\end{figure*}

\begin{figure}[!t]
	\centering
	\includegraphics[trim={0.24cm 0.25cm 0.25cm 0.25cm}, clip=true, width=0.97\linewidth]{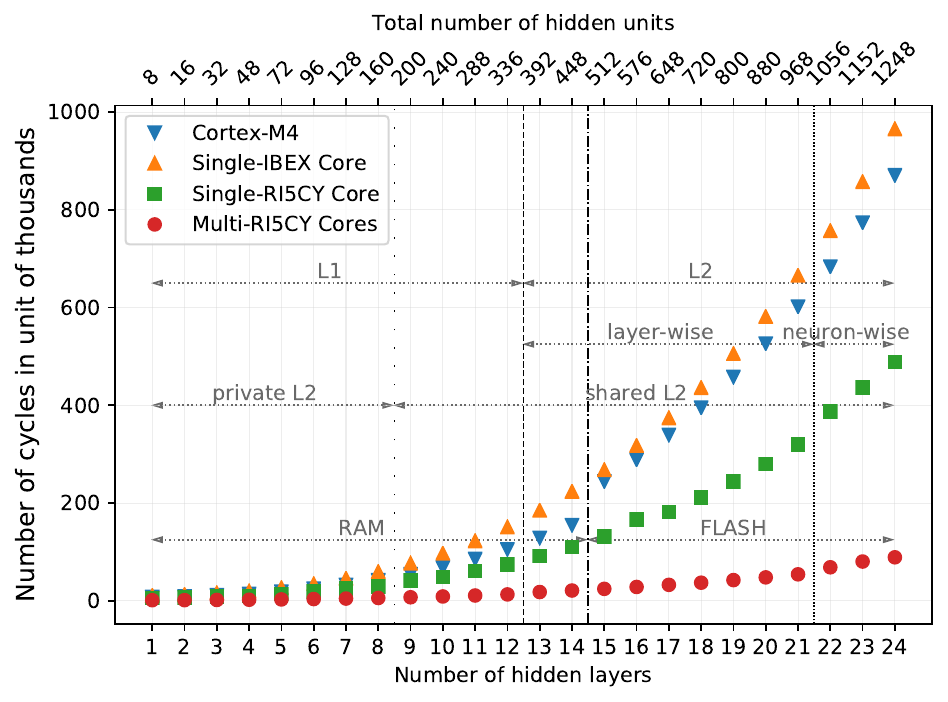}
	\caption{Runtime in number of cycles of whole network with varying number of hidden layers and hidden units.}
	\label{fig:whole_net_num_cycles}
\end{figure}

\begin{figure*}[!t]
    \centering
    \subfloat[]{\includegraphics[trim={0.24cm 0.25cm 0.25cm 0.25cm}, clip=true, width=0.97\columnwidth]{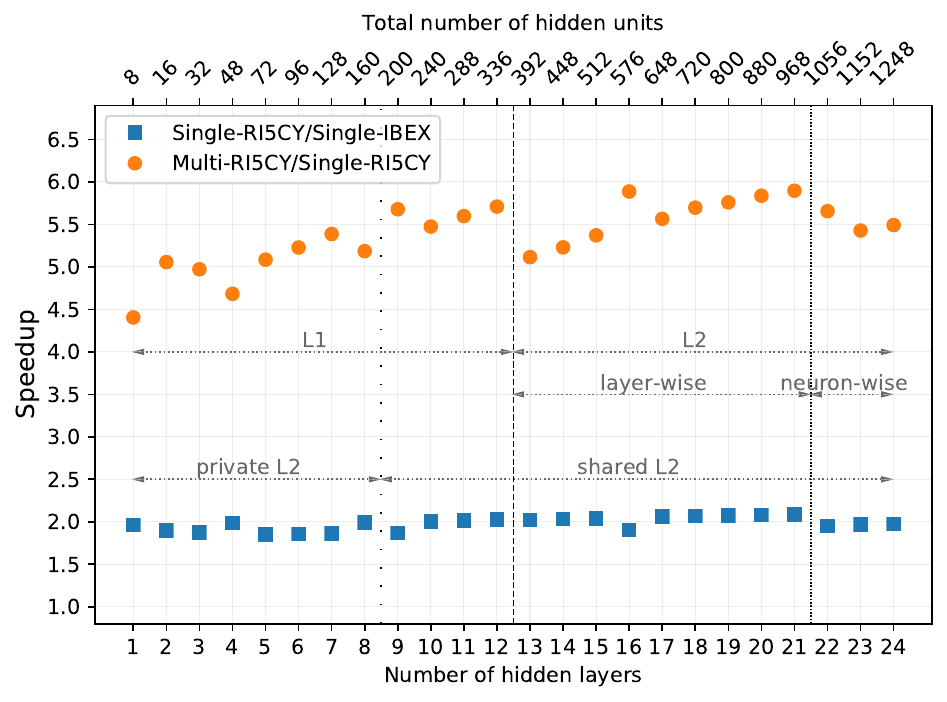}\label{fig:whole_net_pulp_speedup}}
    \hfill
    \subfloat[]{\includegraphics[trim={0.24cm 0.25cm 0.25cm 0.25cm}, clip=true, width=0.97\columnwidth]{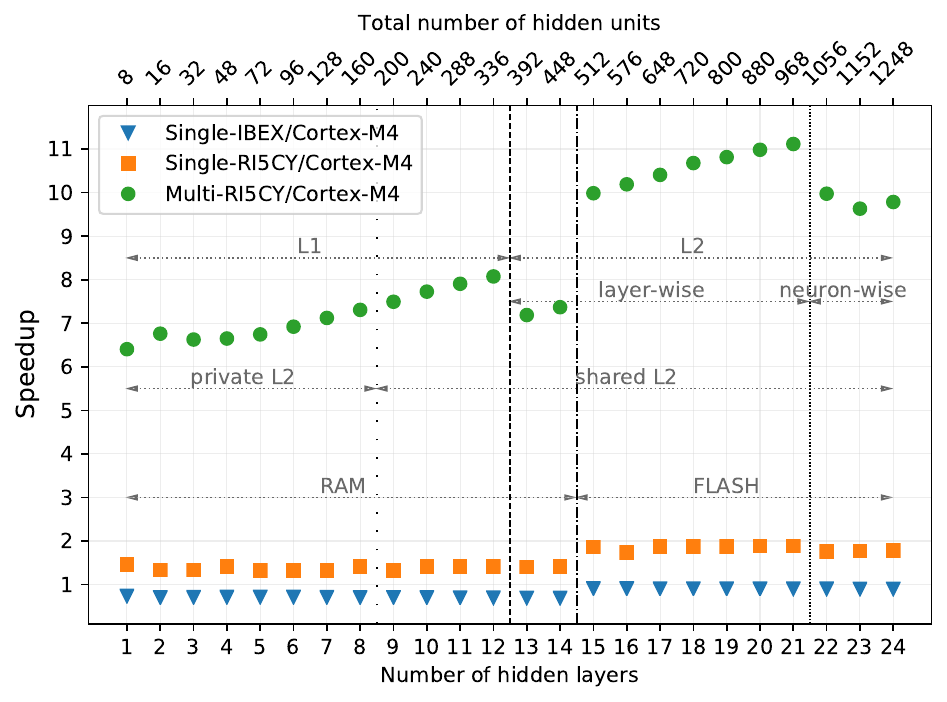}\label{fig:whole_net_speedup}}
    \caption{Speedup measurements of whole network with varying number of hidden layers and hidden units. (a) Speedup measurements on PULP Mr. Wolf, (b) Comparison between PULP Mr. Wolf and ARM Cortex-M4.}
    \label{fig:speedup_end-to-end}
\end{figure*}

We further proceeded with a single-layer performance examination by focusing on the computation of weight matrices without activation functions. \figref{fig:single_layer_arm} shows the runtime in number of cycles of a single layer by varying the number of input and the number of output to the layer executed on STM32L475VG with ARM Cortex-M4 with the fixed-point implementation. The blue grid delimits the case when the layer is too large to fit into RAM and is thus stored in flash memory. ``0.0" indicates that the layer is even too large to be stored in flash memory. \figref{fig:single_layer_fc} shows the same measurements on the FC of Mr. Wolf, i.e., single \ibex core with the basic RM32IM instruction set. The dotted purple grid marks the fact that the layer does not fit into the private L2 memory anymore and is thus stored in the larger shared L2 memory. These two figures represent the reference in number of cycles for the following discussions, which will be led in terms of speedups.

Compared to a single \ibex core, we can see from the results in \figref{fig:single_layer_single_riscy_fc} that we gain up to 2.2$\times$ speedup using a single \riscy core, by adding the custom instruction set extensions, most importantly hardware loops and post-increment loads. The grey dash-dotted grid delimits the results when the layer does not fit into the L1 memory, and neuron-wise double-buffering transfer using DMA unit is applied, i.e., the weights of a single neuron is transferred from the L2 to the L1 memory while the \riscy processor computes the result of the previous neuron. We can see that the speedup for larger input sizes is higher than the speedup for smaller input sizes because the overhead of activating the DMA transfers becomes negligible with longer computation time due to larger input feature vectors. We further measured the speedup of parallel execution with respect to single \riscy core execution. The results in \figref{fig:single_layer_multi_riscy_single_riscy} demonstrate up to 7.7$\times$ speedup with measurements on-board.

Subsequently, we compared the execution on ARM Cortex-M4 and PULP Mr. Wolf. \figref{fig:single_layer_arm_single_riscy} shows the speedup of single \riscy core over ARM Cortex-M4, while \figref{fig:single_layer_arm_multi_riscy} demonstrates the parallel \riscy core speedup over ARM Cortex-M4. The former reaches almost a $2\times$ speedup, whereas the latter achieves up to 13.5$\times$ speedup.

Finally, we measured the performance of entire networks with fixed input and output layers while varying the number of hidden layers and hidden units, as described in \secref{sec:perf_evaluation}. \rebuttal{We used the following criteria to choose the parameter $d$ in \eqref{eq1} and \eqref{eq2}:
\begin{enumerate}
    \item To show at least two measurement points in each memory section, for example between the dashed line and the dash-dotted line in \figref{fig:whole_net_speedup}, i.e. when the network fits into RAM on STM32 with ARM Cortex-M4 and not in L1 of PULP Mr. Wolf.
    \item To include all the possible cases of memory transfer, i.e. both layer-wise transfer and neuron-wise transfer. If d is small, the case of neuron-wise transfer will be absent, because the number of neurons in each layer grows too slowly before the whole network is too big to fit into the largest available memory.
    \item To have the highest energy efficiency with parallelization.
\end{enumerate}
Based on the estimated memory in \eqref{eq0} and running several experiments, $6 < d < 12$ satisfies the criteria 1 and 2 but not the criterion 3. With $d = 8$ all the above mentioned decision policies are met. With this chosen value, we show the performance measurements from relatively small networks with a single hidden layer with only 8 hidden nodes, which can fit into RAM of Cortex-M4 and L1 and private L2 of Mr. Wolf, to relatively large networks with 24 hidden layers with 1248 hidden units, where neuron-wise transfer is required for an optimized execution. With this design, networks larger than 24 hidden layers do not fit into the largest memory of our selected processors. Moreover, since Mr. Wolf has eight parallel cores, networks' layers with multiple of 8 number of neurons exploit the parallelization at its highest efficiency. \figref{fig:whole_net_num_cycles} shows the runtime in number of cycles, and \figref{fig:whole_net_pulp_speedup} and \figref{fig:whole_net_speedup} present the speedup.
}

As can be seen in \figref{fig:whole_net_pulp_speedup}, the network fits into the L1 memory of Mr. Wolf up to 12 hidden layers, i.e., 336 hidden units. Networks larger than 12 hidden layers are stored in the L2 memory and are transferred piece-wise to the L1 memory with DMA transfers in a double-buffering configuration. The largest layer of the networks with between 13 and 21 hidden layers still fits into the L1 memory allowing for layer-wise DMA transfers, i.e., the parameters of a whole layer are transferred at a time. For networks with more than 21 hidden layers, neuron-wise DMA transfers are performed as not all layers fit into the L1 memory even individually. Thus, we transfer the parameters of one neuron at a time. We can see from the plot that when the network size is small, the parallel speedup is lower than in cases where the network size is large. This can be attributed to the overhead of parallelization and is more noticeable for small networks where the amount of computation is relatively small. However, we can still reach around 4.5$\times$ parallel speedup with respect to the single \riscy core for tiny networks such as the one with only one hidden layer comprising 8 hidden neurons. The general tendency of the parallel speedup increases while augmenting the network size.

\figref{fig:whole_net_speedup} demonstrates the comparisons between PULP Mr. Wolf and ARM Cortex-M4. We can see that \ibex core is slightly slower than Cortex-M4 when the latter accesses to RAM, while the performance of Cortex-M4 degrades slightly when the network is too large to fit into RAM, and it has to access the flash memory. In this case, the \ibex core is as fast as the Cortex-M4. The same phenomenon can be observed in the other two figures, i.e., single-core and multi-core speedup of \riscy over Cortex-M4. Specifically, a single \riscy core is almost twice as fast as a Cortex-M4 core as a result of the custom ISA extensions, e.g., hardware loops and post-increment loads. The parallel speedup grows steadily with increasing network size. As expected, we see a speedup drop when the network is too large to be stored in L1 for \riscy cores, while it still fits into the RAM of the Cortex-M4. Contrarily, when the network is stored into flash on the Cortex-M4, the DMA transfer from L2 to L1 in Mr. Wolf offers much more gain with respect to flash access in Cortex-M4, more specifically the speedup reaches up to 11.1$\times$ for the designed network architecture. As mentioned previously, the purpose of this experiment is to show the performance of an entire network starting with very small network sizes, such as the one with only 8 hidden units, and increasing the network size by adding hidden layers and hidden neurons consecutively. With this design, the networks with more than 24 hidden layers do not fit into the largest memory. Nevertheless, the performance of larger networks using neuron-wise DMA transfer is exhaustively presented in figures \ref{fig:single_layer_single_riscy_fc}, \ref{fig:single_layer_multi_riscy_single_riscy}, \ref{fig:single_layer_arm_single_riscy}, and \ref{fig:single_layer_arm_multi_riscy}, and discussed in the first part of this subsection.

\section{Application Showcases} \label{sec:application}

MLPs have been successfully used in a wide range of application
scenarios, such as disease detection~\cite{Das2018}, activity
recognition~\cite{Gaikwad2019}, and brain-machine interface~\cite{Colli-Alfaro2019}. Many studies identified MLPs to be the
best or one of the best algorithms to solve tasks in the IoT domain using wearable
devices~\cite{howcroft2016, Sazonov2015, LeMoyne2016, Can2019}. In this section,
we present three application showcases found in the literature using an MLP
with different network sizes~\cite{Colli-Alfaro2019, howcroft2016, Gaikwad2019}
in order to demonstrate the usability and the power efficiency of our proposed
framework with the supported MCUs. We reproduced the network
architectures and executed the classifications on both Nordic nRF52832 with ARM Cortex-M4 and PULP Mr. Wolf present on InfiniWolf using our proposed framework and measured the runtime and the power consumption.

\subsection{Hand Gesture Recognition} \label{subsec:a-hand_gesture_recognition}
The authors in~\cite{Colli-Alfaro2019} presented a gesture classification method based on a sensor fusion technique using surface electromyography (EMG) and a 9-axes inertial measurement
unit (IMU). The EMG and IMU signals were acquired using Myo Armband~\cite{myoarmband} placed around the forehand. The goal is to recognize 10 hand gestures. 76 time-domain features are extracted from EMG and IMU signals and fed as input to the designed MLP with three hidden layers of 300, 200, and 100 hidden units each, and an output layer of 10 classes. The highest classification accuracy achieved is 85.58\%. As a shorthand, we name it application A.

\subsection{Fall Detection for Elderly People} \label{subsec:b-fall_detection}
We implemented the model proposed in ~\cite{howcroft2016}, named application B, where a combination of pressure sensors in the insoles and accelerometers at the head, pelvis, left, and right shanks of the shoes is proposed to assess fall-risk in elderly people. The authors extracted spatial, temporal, and frequency domain features from the pressure and accelerometer data and explored various sensor combinations and three different machine learning models, i.e., MLPs, naïve Bayesian, and Support Vector Machine. As a result, the best performing model was an MLP with input parameters from pressure sensors and accelerometers at the head, pelvis, and left shank, reaching the best accuracy of 84\%. The network is composed of 117, 20, and 2 nodes, respectively, as input, hidden, and output layers. 

\subsection{Human Activity Classification} \label{subsec:c-human_activity_classification}
The authors in~\cite{Gaikwad2019} proposed an FPGA implementation of MLPs with parallel computation to classify human activity in real-time. The dataset is acquired by a 3-axial accelerometer worn on the waist and classified into five activity classes. Various MLP topologies were investigated, and the final proposed architecture consists of 7 input features extracted using a sliding window, 6 hidden nodes arranged in a single layer, and 5 output nodes, with the best accuracy being 94.6\%. The trained model runs on a Xilinx FPGA with an execution time of 270\,ns and a power consumption of 241\,mW. Here we call it application C for convenience.

\subsection{Experimental Evaluation and Results}
We reproduced the network architectures described above with sigmoidal activation functions using the FANN library and deployed them with our framework on the two processors present on InfiniWolf, i.e., the Nordic nRF52832 with an ARM Cortex-M4 and Mr. Wolf, using our proposed framework. The measurements on nRF52832 are done with the processor running at 64\,MHz and with DC/DC regulator enabled, since it is stated on the datasheet that using the DC/DC regulator will reduce current consumption compared to when using the LDO regulator. The measurements on Mr. Wolf are done considering both the SoC and Cluster domains with the processors running at 100\,MHz since it is shown that at this frequency, the energy efficiency is maximized~\cite{pullini2018}. The measurements are performed using the power analyzer Keysight N6705C with the minimum sampling interval being 0.1024\,ms. 

\begin{figure}[!t]
	\centering
	\includegraphics[trim={0cm 0.3cm 0cm 0.7cm},clip=true, width=\linewidth]{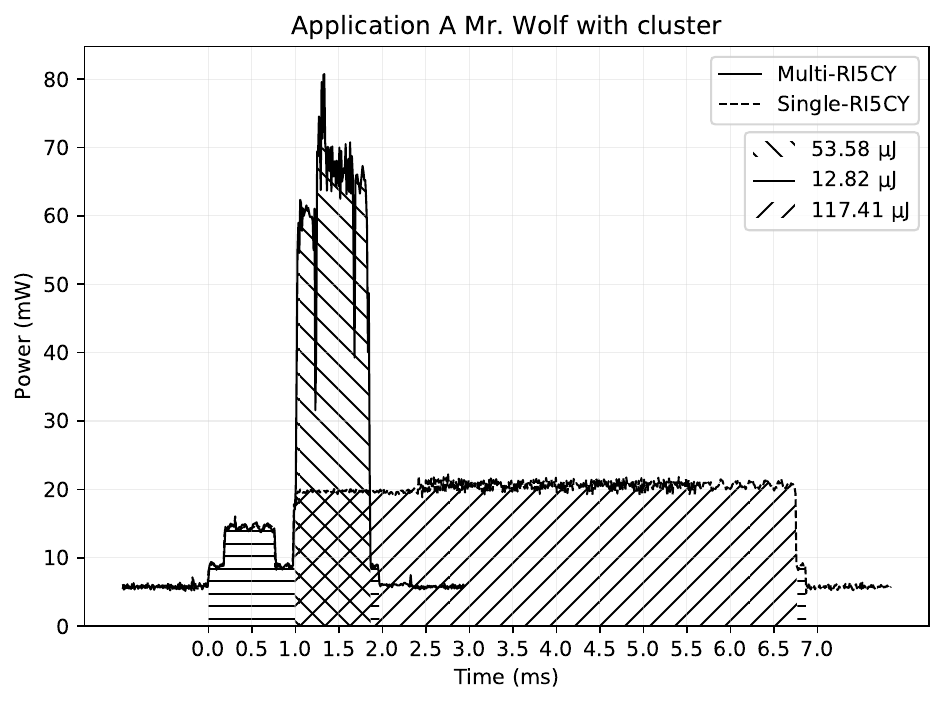}
	\caption{End-to-end power measurements of one classification of application A on Mr. Wolf with \riscy core(s).}
	\label{fig:power_measurements_riscy}
\end{figure}

\tabref{table:application_showcases_runtime_power} summarizes the measured runtime and the average power consumption of the three MLPs used in the three applications. Let us first analyze the application A which has the largest network architecture, i.e., 76-300-200-100-10, yielding 103800 MACs. We can see that the runtime parallel speedup using multi \riscy cores is 7.1$\times$ with respect to a single one, which corresponds to the measurements shown in \figref{fig:single_layer_multi_riscy_single_riscy}. 
For comparisons with \ibex and Cortex-M4, we have to consider additionally the activation, the initialization, and the deactivation of the cluster, that introduce a constant overhead of 1.2\,ms on average, and the DMA transfers of the input data from L2 to L1, which in this case is negligible with 76 inputs (\texttildelow 2.5\,$\upmu$s). \figref{fig:power_measurements_riscy} plots the measured power consumption during the end-to-end execution of a single classification, including the cluster activation/deactivation and DMA transfers. Due to Amdahl's law, the overall runtime speedup of multi \riscy cores is 3.5$\times$ with respect to a single core for one classification. If we do multiple classifications, the constant overhead will become negligible as we increase the number of classifications. Moreover, this overhead can be reduced with improved drivers, which is not the focus of this work. 

Considering the end-to-end performance of a single classification, we obtain 5.7$\times$ and 8.8$\times$ speedup using multi \riscy cores with respect to \ibex and Cortex-M4. These figures grow asymptotically towards 14.3$\times$ and 22$\times$ with an increasing number of MLP classifications per cluster activation. Similar conclusions can be drawn for the energy consumption. The constant overhead is around 13\,$\upmu$J, while the energy consumed for one classification using parallel computation is 54\,$\upmu$J, which is around 2.2$\times$ more energy efficient than the single-core computation. Comparing the end-to-end parallel performance of one classification to Cortex-M4 and \ibex, we save respectively 2.8$\times$ and 1.8$\times$ more energy. These gains increase up to 3.4$\times$ and 2.3$\times$, with a rising number of classifications per cluster activation.

Finally, for very small networks, such as the ones used in applications B and C, the runtime is far below the millisecond range. If the application scenario requires only very few classifications per cluster activation, then the \ibex core is the most energy-efficient one, with a consumption of 2.9\,$\upmu$J and 0.15\,$\upmu$J, respectively for applications B and C. Comparing to the work in~\cite{Gaikwad2019} for application C, the \ibex core is 13.5$\times$ faster in computation time and 434$\times$ more energy efficient than a parallel FPGA implementation. However, if continuous classification is required, which is the case for the vast majority of the IoT applications, then the parallel execution, once again, outperforms in terms of speed and energy efficiency. For example, for one classification in application B, \ibex core consumes 2.86\,$\upmu$J, while the parallel execution consumes 0.67\,$\upmu$J in addition to the constant overhead of 13\,$\upmu$J which is spent only once, then the parallel ultra-low-power approach already pays off when more than 6 classifications are done. If a continuous classification is required, the parallel approach is 4$\times$ as energy-efficient as the single \ibex core.

\begin{table*}[!t]
\caption{Measured Runtime and Average Power Consumption during computation on Application Showcases. \rebuttal{Relative improvements per inference w.r.t. ARM Cortex-M4 implementation in execution time speedup and in energy reduction are reported in parenthesis.}}
\label{table:application_showcases_runtime_power}
\centering
\begin{threeparttable}
\setlength{\tabcolsep}{20pt}
\begin{tabular}{crrrr}
\toprule
\vspace{0.1cm}
    \multirow{2}{*}{App.}    & \multicolumn{1}{c}{nRF52832}         & \multicolumn{3}{c}{-----------------------------\,PULP Mr. Wolf\,-----------------------------}                                \\ 
         & ARM Cortex-M4       & \ibex             & Single-\riscy & Multi-\riscy \\ \midrule
\rowcolor{Gray}
 & 17.6\,ms      & 11.4\,ms \rebuttal{(1.54$\times$)}      & 5.7\,ms\tnote{*} \rebuttal{(3.09$\times$)}      & 0.8\,ms\tnote{*} \rebuttal{(22$\times$)} \\
 \rowcolor{Gray}
 & 10.44\,mW     & 10.75\,mW     & 20.35\,mW     & 61.79\,mW \\
 \rowcolor{Gray}
 \multirow{-3}{*}{A} 
                    & \rebuttal{183.74\,$\upmu$J}     & \rebuttal{122.55\,$\upmu$J (-33.30\%)}     & \rebuttal{116.00\,$\upmu$J (-36.87\%)}            & \rebuttal{49.43\,$\upmu$J (-73.10\%)}      \\
\multirow{3}{*}{B}  & 0.4\,ms       & 0.3\,ms \rebuttal{(1.33$\times$)}       & 0.14\,ms\tnote{*} \tnote{,} \tnote{$\dagger$} \rebuttal{(2.86$\times$)}         & 0.03\,ms\tnote{*} \tnote{,} \tnote{$\dagger$} \rebuttal{(13.33$\times$)} \\
                    & 11.21\,mW     & 9.52\,mW      & 17.54\,mW     & 22.18\,mW \\
                    & \rebuttal{4.48\,$\upmu$J}     & \rebuttal{2.86\,$\upmu$J (-36.16\%)}     & \rebuttal{2.46\,$\upmu$J (-45.09\%)}            & \rebuttal{0.6654\,$\upmu$J (-85.15\%)}      \\
                  \rowcolor{Gray}
  & 0.03\,ms\tnote{$\dagger$}          & 0.02\,ms\tnote{$\dagger$} \rebuttal{(1.5$\times$)}  & 0.01\,ms\tnote{*} \tnote{,} \tnote{$\dagger$} \rebuttal{(3$\times$)}          & 0.004\,ms\tnote{*} \tnote{,} \tnote{$\dagger$} \rebuttal{(7.5$\times$)} \\
  \rowcolor{Gray}
  & 9.74\,mW           & 7.31\,mW   & 16.91\,mW          & 17.17\,mW \\
  \rowcolor{Gray}
  \multirow{-2}{*}{C}
                    
                    & \rebuttal{0.2922\,$\upmu$J}     & \rebuttal{0.1462\,$\upmu$J (-49.97\%)}     & \rebuttal{0.1691\,$\upmu$J (-42.13\%)}            & \rebuttal{0.06868\,$\upmu$J (-76.50\%)}      \\
                    \bottomrule
\end{tabular}
\begin{tablenotes}\footnotesize 
\item[*] In addition around 1\texttildelow 1.3\,ms for cluster activation, initialization, and deactivation with an average power consumption of 11.88\,mW.
\item[$\dagger$] The runtime is below or very close to the precision of the measuring instrument, hence it is calculated from the measured number of cycles.
\end{tablenotes}
\end{threeparttable}
\end{table*}

\section{Discussion} \label{sec:discussion}
\rebuttal{
In this work we presented FANN-on-MCU, a toolkit to implement efficient ANNs on both ARM Cortex-M and RISC-V-based PULP processors. We also analyzed the performance of our toolkit by varying the number of hidden layers and hidden neurons. We demonstrated that the parallelization over eight cores offers up to 7.1$\times$ speedup with respect to a single core. However, the usage of more cores means more power consumption, as can be noticed in \figref{fig:power_measurements_riscy}. In a future work, the trade-off between the number of active cores, i.e. power consumption, and the parallel speedup is to be analyzed. We further analyzed into more details three different networks used in three application showcases with runtime and power measurements. However, the limitation in this approach is that the data acquisition and preprocessing of data are not considered. In future work, we are planning to implement the full chain of data acquisition and processing to fully evaluate the complete system.
}

\section{Conclusion} \label{sec:conclusions}
Artificial intelligence and machine learning for low-power IoT devices that host MCUs are key technologies for near-sensor data analytics and decision making. Multi-layer neural networks are showing incredible performance in terms of accuracy in many applications. However, very few implementations and measurement results exist for low-power MCUs. We have presented FANN-on-MCU, a framework to facilitate deployment of optimized neural networks trained using the open-source FANN library. Our framework supports not only the very popular ARM Cortex-M series MCUs, but also the RISC-V-based Parallel Ultra-Low Power (PULP) processors, both with and without a floating-point unit. We have further shown performance comparisons of neural network inference between the two classes of processors and \rebuttal{analyzed the parallel speedups and degradations due to parallelization overhead and memory transfers. Experimental results have shown that a parallel implementation on PULP-based Mr. Wolf can reach up to 7.1$\times$ speedup with respect to a single core implementation, while it outperforms an ARM Cortex-M4 up to 13.5$\times$. Moreover, we presented three different use-cases to evaluate the energy efficiency of our framework.} We have demonstrated that ANNs are suitable for deployment on ultra-low-power MCUs in terms of memory usage, compute time, and energy consumption. Specifically, experimental measurements have shown that using FANN-on-MCU, \rebuttal{a relatively big network with 103800 multiply-accumulate operations can be executed within 17.6\,ms using 183\,$\upmu$J on a Nordic nRF52832 MCU with an ARM Cortex-M4, whilst the parallel implementation on Mr. Wolf with 8 RISC-V-based \riscy cores executes the same inference in less than 1\,ms consuming around 50\,$\upmu$J. More generally, a parallel implementation on Mr. Wolf offers on average an 80\% energy reduction and a 14$\times$ speedup in inference compared to an ARM Cortex-M4 implementation.} Finally, the framework is released open-source and ready to be used to deploy neural networks for applications on low-power embedded systems. 

\bibliographystyle{IEEEtran}
\bibliography{bstctl,ref,bib}

\begin{IEEEbiography}[{\includegraphics[width=1in,height=1.25in,clip,keepaspectratio]{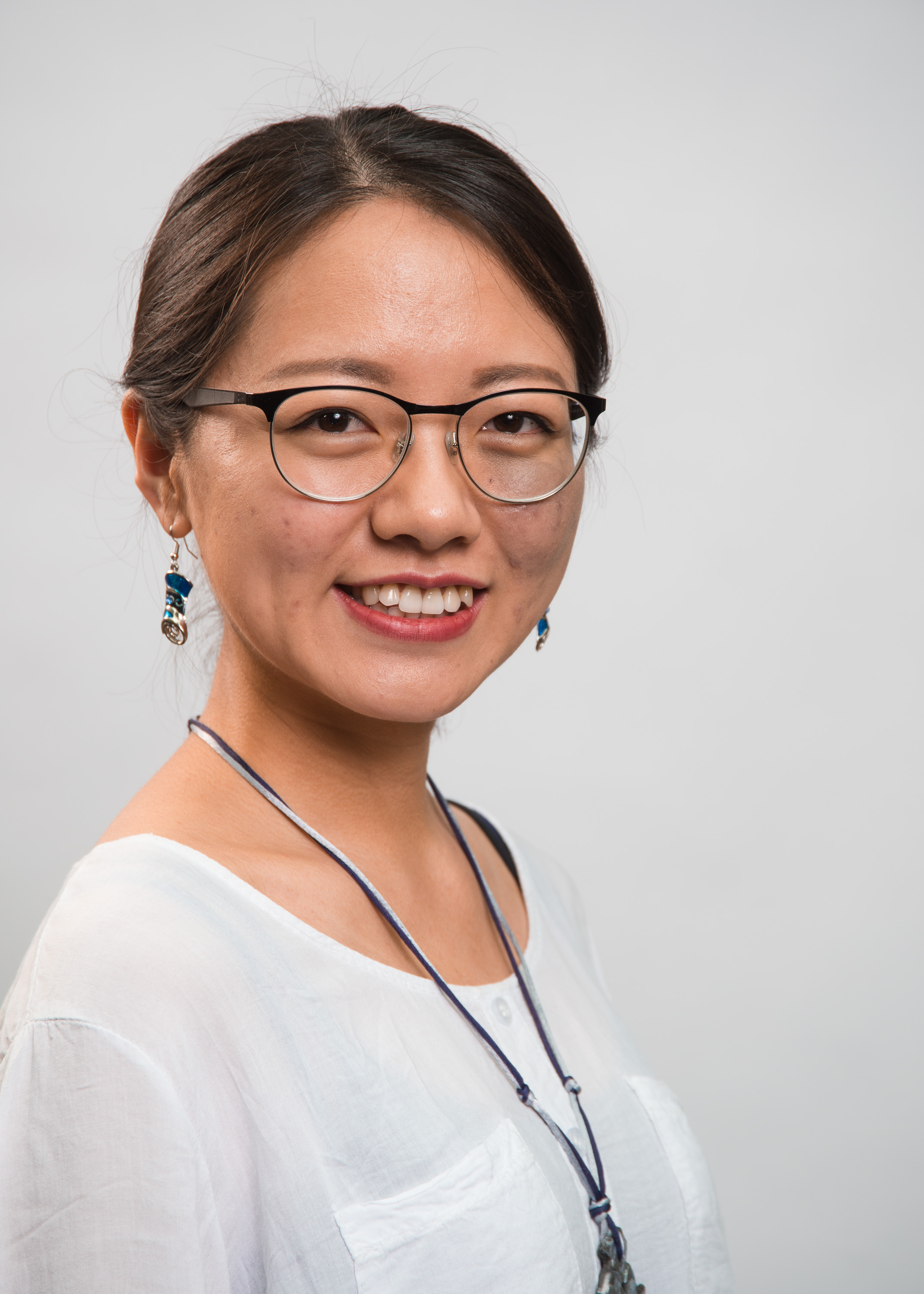}}]{Xiaying Wang}
received her B.Sc. and M.Sc. degrees in biomedical engineering from Politecnico di Milano, Italy and ETH Zürich, Switzerland in 2016 and 2018, respectively. She is currently pursuing a Ph.D. degree at the Integrated Systems Laboratory at ETH Zürich. Her research interests include biosignal processing, low power embedded systems, energy-efficient smart sensors and machine learning on microcontrollers. She received the excellent paper award at the IEEE Healthcom conference in 2018 and she won the Ph.D. Fellowship funded by Swiss Data Science Center in 2019.
\end{IEEEbiography}

\begin{IEEEbiography}[{\includegraphics[width=1in,height=1.25in,clip,keepaspectratio]{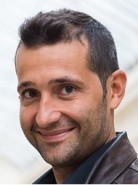}}]{Michele Magno}
received the masters’ and Ph.D. degrees in electronic engineering from the University of Bologna, Bologna, Italy, in 2004 and 2010, respectively. He is currently a Senior Researcher and lecturer with ETH Zürich, Switzerland. He has authored more than 150 papers in international journals and conferences, few of them awarded as best paper. His current research interests include wireless sensor networks, wearable devices, energy harvesting, low power management techniques, and extension of the lifetime of batteries-operating devices. 
\end{IEEEbiography}

\begin{IEEEbiography}[{\includegraphics[width=1in,height=1.25in,clip,keepaspectratio]{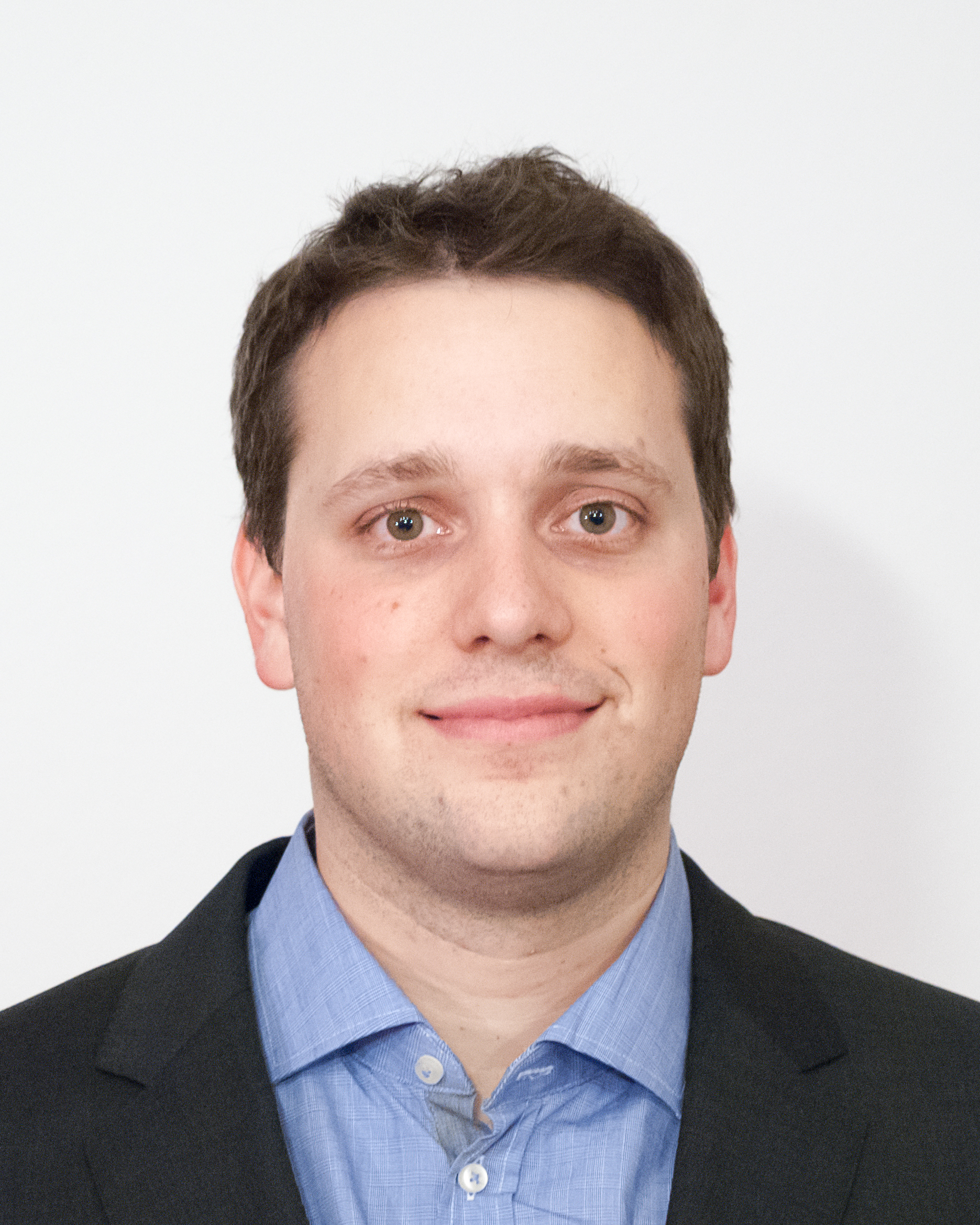}}]{Lukas Cavigelli}
received the B.Sc., M.Sc., and Ph.D. degree in electrical engineering and information technology from ETH Zürich, Zürich, Switzerland in 2012, 2014 and 2019, respectively. He has since been a postdoctoral researcher with ETH Zürich. His research interests include deep learning, computer vision, embedded systems, and low-power integrated circuit design. He has received the best paper award at the VLSI-SoC and the ICDSC conferences in 2013 and 2017, the best student paper award at the Security+Defense conference in 2016, and the Donald O. Pederson best paper award (IEEE TCAD) in 2019.
\end{IEEEbiography}

\begin{IEEEbiography}[{\includegraphics[width=1in,height=1.25in,clip,keepaspectratio]{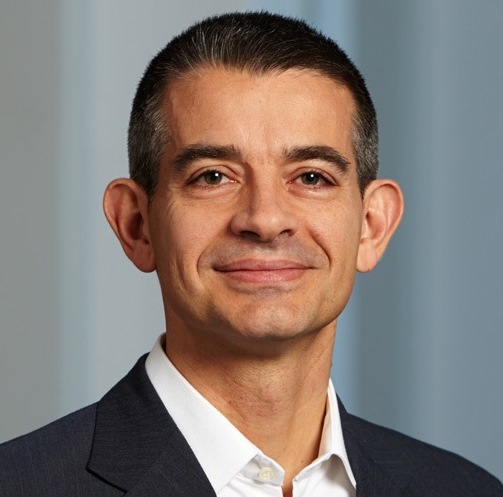}}]{Luca Benini}
is the Chair of Digital Circuits and Systems at ETH Zürich and a Full Professor at the University of Bologna. He has served as Chief Architect for the Platform2012 in STMicroelectronics, Grenoble. Dr. Benini’s research interests are in energy-efficient system and multi-core SoC design. He is also active in the area of energy-efficient smart sensors and sensor networks. He has published more than 1’000 papers in peer-reviewed international journals and conferences, four books and several book chapters. He is a Fellow of the ACM and of the IEEE and a member of the Academia Europaea.
\end{IEEEbiography}

\end{document}